\documentclass[pdflatex,sn-mathphys-num]{sn-jnl}

\usepackage{amsmath,amssymb,amsfonts}
\usepackage{xcolor}
\usepackage{booktabs}
\usepackage{subcaption}
\usepackage{cleveref}
\usepackage{rotating}
\usepackage{colortbl}
\usepackage{makecell}
\usepackage{graphicx}
\usepackage{multirow}
\usepackage{algorithm}
\usepackage{algorithmicx}
\usepackage{algpseudocode}

\begin{document}

\title[Evolutionary Giant Tour for CVRP using NSE and ML Heuristic]{Evolutionary Giant Tour approach for CVRP using Node Shift Encoding and Machine Learning repair heuristic}

\author[]{\fnm{Souad} \sur{Abdoune}}\email{s.abdoune@univ-boumerdes.dz}
	
\author*[]{\fnm{Menouar} \sur{Boulif*}}\email{boumen7@gmail.com}
	
\affil*[]{\orgdiv{LIMOSE Laboratory}, \orgname{University M'hamed Bougara of Boumerdes}, \orgaddress{\postcode{35000}, \state{Boumerdes}, \country{Algeria}}}

\abstract{The Capacitated Vehicle Routing Problem (CVRP) remains a central concern in logistics research, as route optimisation under rigid capacity and fleet restrictions directly affects operational performance. Genetic Algorithms are widely used for this problem, yet their effectiveness depends heavily on the choice of solution encoding and on the mechanisms used to preserve feasibility during the search. This paper proposes a sequential framework that separates the evolutionary process from route construction. We adapt Node Shift Encoding (NSE) to the CVRP giant-tour setting and benchmark it against Path Representation and Double Chromosome. We further propose an unsupervised machine-learning repair heuristic that restores fleet feasibility after an unconstrained linear split, and compare this pipeline against fleet-constrained split taken from the literature. Computational tests on standard benchmark instances show that NSE consistently outperforms the other encodings in solution quality and feasibility. The repair heuristic restores feasibility in approximately 72\% of cases, and the Linear-Split-ML-Heuristic pipeline proves more robust than the constrained split, delivering lower costs and higher feasibility rates.}
	
	
\maketitle

\section{Introduction}\label{sect:Intro}

Vehicle Routing Problems (VRPs) underpin a wide range of physical distribution systems, from last-mile delivery to industrial supply chains \cite{Toth2014, Abdoune2025towards}. Because routing decisions directly affect fuel consumption, fleet utilization, and service times, even modest improvements in tour planning can translate into substantial operational savings and reduced environmental impact \cite{Toth2014}. Structurally, the VRP generalizes the classical Travelling Salesman Problem (TSP): whereas the TSP asks for a single minimum-cost circuit visiting every city exactly once before returning to the depot \cite{Applegate2011}, the VRP must coordinate multiple routes that depart from a central depot, serve disjoint customer subsets, and respect capacity, time, or fleet-size limits \cite{Dantzig1959}. This added layer of combinatorial complexity places the VRP in the NP-hard class \cite{Lenstra1981}, and its capacitated variant--the Capacitated Vehicle Routing Problem (CVRP)--has become a standard benchmark for routing methodologies \cite{Laporte2009}.

Given this computational hardness, exact methods such as branch-and-cut or mathematical programming guarantee optimality for relatively small or modest instances \cite{Ralphs2003, Baldacci2008}. For the larger problem sizes encountered in practice, meta-heuristics--including Genetic Algorithms (GAs)--are widely used because they can deliver high-quality solutions within strict time budgets \cite{Vidal2013}.

A prominent and effective strategy for solving the CVRP is the \emph{giant tour} paradigm \cite{Beasley1983}. In this two-phase approach, the algorithm first treats the instance as a single-vehicle TSP and evolves an unconstrained Hamiltonian cycle visiting all customers. A splitting procedure then partitions this global sequence into capacity-feasible vehicle routes \cite{Prins2004, Vidal2016}. This separation of sequencing from partitioning allows the search to exploit strong TSP operators while delegating route construction to an efficient routine \cite{Prins2004, Vidal2016}.

Yet this sequential decomposition introduces two practical difficulties for evolutionary approaches. First, the final route quality depends strongly on how the giant tour is encoded. Traditional Path Representation (PR) is simple, but standard crossover operators can disrupt useful sub-sequences, which in turn may reduce the effectiveness of the subsequent split \cite{Larranaga1999}. alternative encoding schemes are introduced to solve the problem such as: The Double Chromosome (DC) \cite{Riazi2019} and Node-Shift Encoding \cite{Boulif2024}. Second, when the splitting phase is applied without explicitly enforcing a fleet limit, the resulting number of routes may exceed the available fleet size \(K\). Existing approaches address this either by incorporating fleet-size constraints directly in the split procedure \cite{Vidal2016} or by embedding penalty terms within the evolutionary loop \cite{Vidal2013}, both of which add constraint-management overhead to the GA.

To address these limitations, we propose a sequential meta-heuristic framework and conduct two complementary comparative studies:
\begin{enumerate}
\item \textbf{Encoding comparison.} We use the Node Shift Encoding (NSE), originally developed for the TSP \cite{Boulif2024}, to the CVRP giant-tour setting and benchmark it against PR and DC. Because NSE encodes relative node shifts rather than absolute positions, it is designed to preserve promising structural patterns during crossover and mutation operations.

\item \textbf{Splitting strategy comparison.} We introduce an unsupervised machine-learning repair heuristic as a standalone post-processing layer. When the standard split \cite{Prins2004, Vidal2016} produces more than \(K\) routes, the heuristic embeds the distance matrix into a two-dimensional Euclidean space using distance-geometry principles \cite{Alencar2015} and reallocates surplus customers to existing routes using centroid guidance and capacity-aware insertion. This restores fleet feasibility without re-entering the GA loop. We compare this \emph{linear split + ML repair} pipeline against the fleet-size constrained splitting method of Vidal \cite{Vidal2016}, which enforces \(K\) directly inside the dynamic-programming recurrence.
\end{enumerate}

Computational experiments on 24 standard Augerat benchmark instances \cite{Augerat1995} indicate that, within the sequential framework, NSE achieved better solution quality than PR and DC. Moreover, the linear split combined with the ML repair heuristic is competitive with Vidal’s constrained split and shows improvements in feasibility rate, solution quality, and convergence speed.

The remainder of this paper is organized as follows: Section \ref{sec:related} reviews the related works. Section \ref{sect:PM} presents the formal mathematical model of the CVRP. Section \ref{sect:approach} details the architecture of our proposed approach, including the sequential GA framework and the ML-based restoration heuristic. Section \ref{sect:Exper} outlines the experimental design. Afterwards, sections \ref{sect:Res} and  \ref{sect:diss} present the computational results and  provide a comprehensive discussion, respectively. Finally, Section \ref{sect: conc} concludes the paper and outlines promising directions for future research.

\section{Related works}
\label{sec:related}

The Capacitated Vehicle Routing Problem has motivated extensive research on two-phase solution frameworks that separate route sequencing from route partitioning. This section reviews the evolution of the giant-tour paradigm, the genetic algorithms employed to construct these tours, the splitting procedures that partition them, and the fleet-size enforcement strategies that remain problematic. We conclude by identifying the specific gaps that our work  addresses.

\subsection{Giant-Tour paradigm}
\label{subsec:gianttour}

The \emph{giant tour} paradigm offers a conceptually elegant reduction of the multi-vehicle CVRP to a single-vehicle sequencing problem. Rather than manipulating multiple routes simultaneously, the approach first constructs one comprehensive Hamiltonian cycle visiting all customers, then partitions this global sequence into capacity-feasible vehicle routes. This \emph{Order-first, Split-second} philosophy was originally formalized by Beasley~\cite{Beasley1983} under the label \emph{Route-first, Cluster-second}. Beasley modeled the partitioning phase as a shortest-path problem on an auxiliary directed acyclic graph (DAG), where arc weights encode sub-route costs and dynamic programming yields the optimal partition in polynomial time.

The paradigm gained substantial traction when Prins~\cite{Prins2004} integrated it with a Genetic Algorithm, demonstrating that a GA could evolve high-quality giant tours while a dedicated splitter handled the capacity consideration. This separation of concerns proved highly scalable and stimulated numerous hybrid evolutionary metaheuristics, surveyed comprehensively by Prins et al.~\cite{Prins2014}. The core insight is that the GA can exploit powerful TSP methods and simple permutation-based encodings, while the splitter guarantees feasibility.

\subsection{Genetic Algorithms for Giant-Tour construction}
\label{subsec:gaga}

Genetic Algorithms have become the metaheuristic of choice for giant-tour construction due to their population-based exploration and straightforward adaptation to permutation spaces~\cite{Holland1975, Holland1992}. The effectiveness of the evolutionary process depends critically on how solutions are encoded.

The dominant encoding is the Path Representation (PR), which represents a giant tour as a direct permutation of customer indices. However, standard crossover and mutation operators applied to PR frequently produce invalid offspring: duplicate customer visits and omitted nodes violate the Hamiltonian cycle constraint.

Multiple corrective strategies have been proposed including: death penalties, reinitialization, or complex restoration method are introduced to restore feasibility.

Two important research directions emerged to resolve this problem. The first direction focused on designing specialized genetic operators tailored to permutation spaces. Operators such as those of \cite{Davis1985, Oliver1987} preserve solution feasibility through complex index mappings, cycle detection, or segment reordering. However, these operations require $O(n^2)$ to $O(n^3)$ time per application, becoming prohibitive within population-based metaheuristics for large CVRP instances. The second direction proposes two-vector representations that separate the genetic material from the decoded solution, notably the Double Chromosome (DC) \cite{Riazi2019} scheme and the Node Shift Encoding (NSE) \cite{Boulif2024}, both originally developed for the TSP. Riazi \cite{Riazi2019} proved that DC outperforms PR, and Boulif \cite{Boulif2024} subsequently demonstrated that NSE outperforms both DC and PR. Motivated by these findings, this paper tests the effectiveness of NSE in the CVRP setting by benchmarking it against DC and PR within a sequential giant-tour framework.

\subsection{Splitting procedures}
\label{subsec:splitting}

Once a giant tour is constructed, the splitting phase must partition it into vehicle routes respecting capacity constraints. The classical splitter, derived from Beasley's DAG formulation, solves this in $O(n^2)$ using dynamic programming~\cite{Beasley1983, Prins2004}. Vidal~\cite{Vidal2016} achieved a significant advance by reducing this to $O(n)$ through exploitation of the Monge property \cite{hoffman1963simple}, maintaining a double-ended queue of candidate paths and pruning dominated alternatives. This linear-time routine removed the splitter as a computational bottleneck and enabled population-based metaheuristics to tackle large-scale instances without algorithmic degradation.

However, these procedures enforce capacity only; they ignore any fleet-size limit $K$ imposed by the problem instance. When the basic split returns more than $K$ routes, the solution is infeasible. Vidal~\cite{Vidal2016} addressed this by extending the dynamic-programming state: $V(k,j)$ denotes the minimum cost to serve the first $j$ customers using exactly $k$ vehicles. Its success depends entirely on the ordering of customers within the giant tour. If the sequence contains no valid cut points that yield exactly $K$ capacity-feasible segments, the dynamic program exhausts all possibilities and returns $V(K,n-1) = \infty$. While instance complexity (tight capacity-to-demand ratios, heterogeneous demand distributions) certainly raises the baseline difficulty, our experiments indicate that the giant tour's ordering is the decisive factor: the same instance, under different GA-generated permutations, frequently transitions from splittable to unsplittable. Since the GA optimizes tour length without regard to partitionability, it may converge to orderings that are inherently unsplittable under the fleet limit regardless of instance complexity. This structural fragility--decoupling the GA's objective from the splitter's feasibility requirements--motivates our post-processing heuristic.

\subsection{Contribution}
\label{subsec:contribution}

We address these limitations through two principal contributions. First, we use Node Shift Encoding to the CVRP giant-tour setting and compare it against Path Representation and Double Chromosome, quantifying how encoding choice affects solution quality and post-split fleet feasibility. Second, we introduce an unsupervised machine-learning repair heuristic triggered exclusively after splitting: when \textit{Linear Split} \cite{Vidal2016} produces excess routes, the heuristic embeds the distance matrix into two-dimensional Euclidean space via distance-geometry principles~\cite{Alencar2015}, computes spatial centroids of surviving routes, and reallocates surplus customers through capacity-aware insertion.

\section{Problem statement}
\label{sect:PM}

The Capacitated Vehicle Routing Problem (CVRP) is a classical combinatorial optimization problem extensively studied in the operations research literature \cite{Toth2014, Laporte2009, Abdoune2025towards}. In what follows, we assume the following hypotheses:

\begin{enumerate}
    \item A single central depot (node 1) and each route starts and ends at the depot.
    \item An homogeneous fleet of identical vehicles, each with limited capacity.
    \item Customer demands are static and known in advance.
    \item Each customer is visited exactly once by exactly one vehicle, and must be fully served (no split delivery).
\end{enumerate}

Under these assumptions, the CVRP can be modeled on a complete undirected graph $G=(V, A)$, where $V = \{1, ..., n\}$ represents the set of nodes, including a depot (node $1$) and $n-1$ customers (nodes $2, ..., n$), and $A$ denotes the set of directed arcs connecting pairs of nodes. Each customer $i \in \{2, ..., n\}$ has a non-negative demand $q_i$, and each vehicle has a maximum capacity $Q$ \cite{Toth2014, Golden2008}. The cost (or distance) of traveling from node $i$ to node $j$ is denoted by $c_{ij}$.

Therefore, solving the CVRP consists of determining a set of at most $K$ vehicle routes that minimizes the total travel cost, while respecting the aforementioned hypotheses. This problem was first introduced by \cite{Dantzig1959} and has since become a cornerstone of distribution and logistics planning \cite{Laporte2009}.

By constructing over the graph $G=(V,A)$, let the binary decision variable be defined as:
\[
x_{ijk} =
\begin{cases}
    1 & \text{if vehicle } k \text{ travels directly from node } i \text{ to node } j, \\
    0 & \text{otherwise},
\end{cases}
\]
for all $i, j \in V$, $k \in \{1, ..., K\}$.

Over this decision variable, we define a mathematical program using the following parameters:
\begin{itemize}
    \item $K$: number of available vehicles;
    \item $Q$: vehicle capacity;
    \item $q_i$: demand of customer $i$ ($q_1 = 0$ for the depot);
    \item $c_{ij}$: cost of traveling from node $i$ to node $j$.
\end{itemize}

The objective function $Z$ is then defined by:
\begin{equation}\label{eq:obj}
\min Z = \sum_{k=1}^{K} \sum_{i=1}^{n} \sum_{j=1}^{n} c_{ij} \cdot x_{ijk}
\end{equation}

Subject to the constraints:

\begin{itemize}
    \item \textbf{Vehicle capacity:} The total demand of customers served by each vehicle must not exceed its capacity \cite{Toth2014}.
    \begin{equation} \label{eq:capacity}
    \sum_{i=1}^{n} \sum_{j=2}^{n} q_j \cdot x_{ijk} \leq Q \qquad \forall k \in \{1, ..., K\}
    \end{equation}
    
    \item \textbf{Degree constraints:} Each customer is visited exactly once \cite{Laporte2009, Toth2014}.
    \begin{equation}\label{eq:enter}
    \sum_{i=1}^{n} \sum_{k=1}^{K} x_{ijk} = 1 \qquad \forall j \in \{2, ..., n\}
    \end{equation}
    \begin{equation} \label{eq:leave}
    \sum_{j=1}^{n} \sum_{k=1}^{K} x_{ijk} = 1 \qquad \forall i \in \{2, ..., n\}
    \end{equation}
    
    \item \textbf{Depot constraints:} Each vehicle departs from and returns to the depot at most once \cite{Golden2008}.
    \begin{equation} \label{eq:depart}
    \sum_{j=2}^{n} x_{1jk} = 1 \qquad \forall k \in \{1, ..., K\}
    \end{equation}
    \begin{equation} \label{eq:return}
    \sum_{i=2}^{n} x_{i1k} = 1 \qquad \forall k \in \{1, ..., K\}
    \end{equation}
    
    \item \textbf{Flow conservation:} For every customer node visited by vehicle $k$, the number of entries equals the number of exits \cite{Toth2014}.
    \begin{equation} \label{eq:flow}
    \sum_{j=1}^{n} x_{ijk} = \sum_{j=1}^{n} x_{jik} \qquad \forall i \in \{2, ..., n\},\ \forall k \in \{1, ..., K\}
    \end{equation}
    
    \item \textbf{Subtour elimination:} To prevent disconnected routes, we introduce auxiliary variables $u_{ik}$ representing the cumulative load of vehicle $k$ upon arriving at customer $i$ \cite{Miller1960, Toth2014}:
    \begin{equation} \label{eq:subtour1}
    u_{ik} - u_{jk} + Q \cdot x_{ijk} \leq Q - q_j \qquad \forall i, j \in \{2, ..., n\},\ i \neq j,\ \forall k \in \{1, ..., K\}
    \end{equation}
    \begin{equation} \label{eq:subtour2}
    q_i \leq u_{ik} \leq Q \qquad \forall i \in \{2, ..., n\},\ \forall k \in \{1, ..., K\}
    \end{equation}
    
    \item \textbf{Binary nature of decision variables:}
    \begin{equation} \label{eq:binary}
    x_{ijk} \in \{0, 1\} \qquad \forall i, j \in \{1, ..., n\},\ \forall k \in \{1, ..., K\}
    \end{equation}   
\end{itemize}

\section{Proposed methodology}
\label{sect:approach}
\begin{figure}[ht!]
	\centering
	\includegraphics[width=0.9\linewidth]{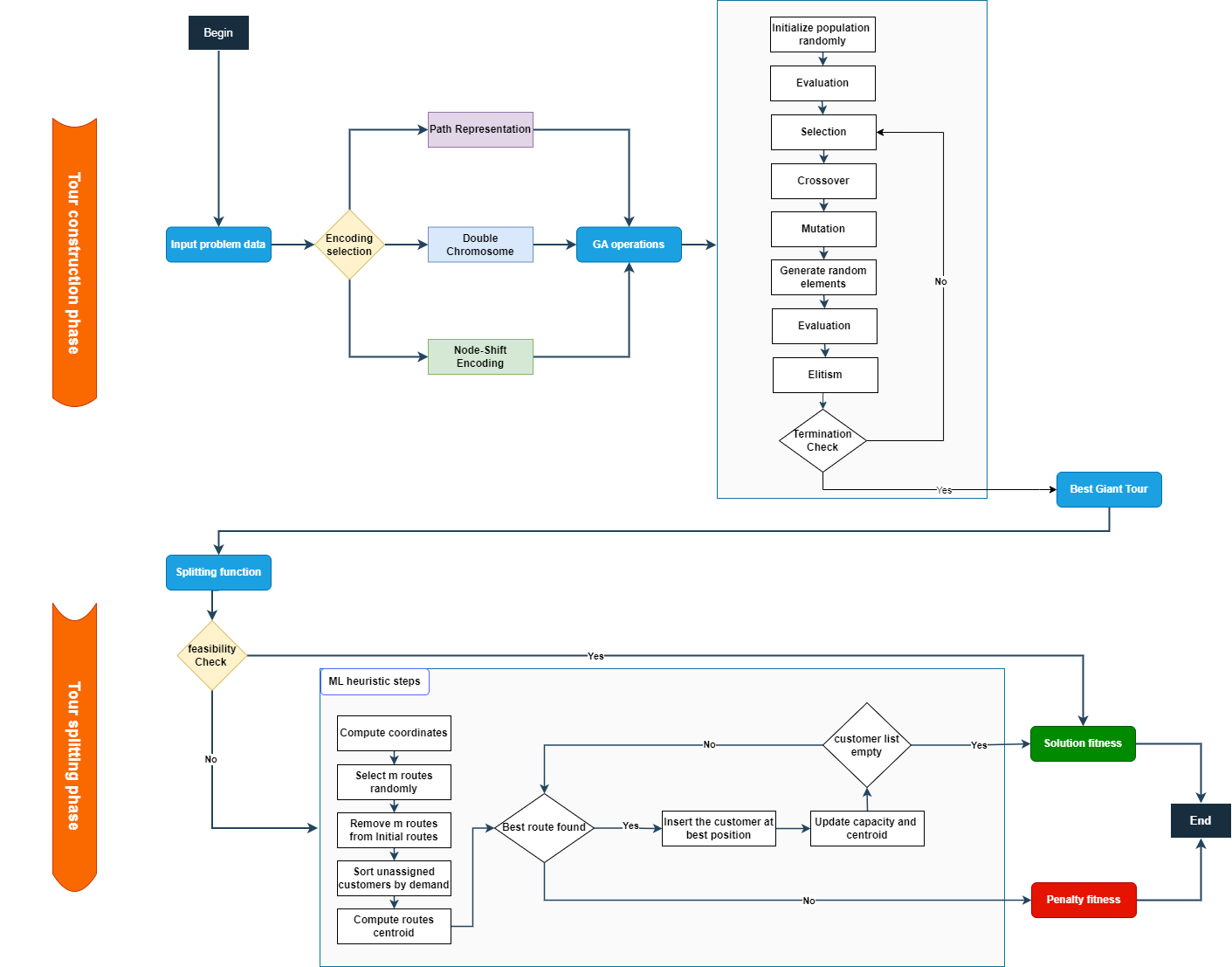}
	\caption{Flowchart of the proposed meta-heuristic resolution framework}
	\label{Fig_flow_chart}
\end{figure}

Our methodology treats the Capacitated Vehicle Routing Problem as an extended sequencing challenge. By temporarily relaxing vehicle capacity and fleet size constraints, the problem is transformed into a single-vehicle Traveling Salesman Problem. This relaxation facilitates the construction of a \emph{giant tour}, a comprehensive Hamiltonian cycle visiting all customers exactly once. Once this global sequence is established, it is systematically partitioned into a set of viable, capacity-constrained routes. The macro architecture driving our unified global solver, from chromosome parsing to the machine learning coordinate repair layer, is schematically detailed in the flowchart of Figure~\ref{Fig_flow_chart}. 

To present this framework clearly, this section is divided into three operational steps: giant tour construction, tour splitting, and the proposed machine-learning repair heuristic.

\subsection{Giant Tour construction} 
We employ a Genetic Algorithm to construct and optimize the giant tour. The GA's inherent ability to balance exploration and exploitation makes it well suited for traversing the vast search space of the CVRP.

\subsubsection{Population representation}
In routing optimization, the representation scheme dictates the design and efficiency of genetic operators. In this work, we introduce the application of Node Shift Encoding to the CVRP context and conduct a rigorous comparative analysis against the classic Path and Double Chromosome representations.
	
\begin{itemize}
   \item \textbf{Path Representation (PR).} This widely adopted encoding represents a tour as a permutation vector of length $n$, where each element corresponds to a customer index in the exact order of visitation. The return to the starting depot is implicit and omitted from the vector representation \cite{Larranaga1999}. While intuitive, standard genetic operators often disrupt the permutation structure, necessitating computational repair mechanisms to fix duplicate or missing nodes. 
  
   \item \textbf{Double Chromosome encoding (DC):} is an encoding strategy that utilizes two separate chromosomes: the map and the guide chromosomes. The map chromosome defines a reference tour, encoding a sequence of customers in a specific order. The guide chromosome, on the other hand, consists of pairs of customer indices that dictate swap operations to be performed on the map chromosome. By applying these swap instructions, the reference tour is modified, leading to the generation
   of a new customer sequence\cite{Riazi2019}.
  
   \item \textbf{Node Shift Encoding (NSE):} This approach utilizes a dual-vector chromosome designed to provide an alternative solution mapping mechanism. The first vector denotes a fixed reference tour of length $n$, representing the total number of customers. The second vector consists of $n-1$ elements containing integers within the range $[0, n-2]$, which represent the positional shift values for each respective customer relative to the reference sequence.
      
   The shifting mechanism is executed sequentially from left to right, where each shift operates dynamically on the current state of the sequence. Modular arithmetic is applied to handle circular movements when a shift value exceeds the remaining sequence length. In NSE, the position of the first element in the reference tour remains structurally anchored, while the relative positions of the remaining $n-1$ customers are determined by decoding the second vector \cite{Boulif2024, Abdoune2026}.    
\end{itemize}
		
\subsubsection{Evaluation}
During this phase, the biological fitness of each individual within the population is evaluated. The primary objective function $Z(s)$ minimizes the total travelled distance of solution $s$. Since standard genetic algorithms are structurally framed as maximization procedures, the raw objective value is transformed into a fitness metric as follows:				
\begin{equation}
\text{Fitness}(s) = B - Z(s)
\label{eq:fitness}
\end{equation}
where $Z(s)$ represents the total distance of solution $s$, and $B$ denotes an upper bound of the objective function. This transformation guarantees that solutions yielding shorter total travel distances map to higher fitness values.

To ensure a tight and instance-specific upper bound, $B$ is computed by sorting all scalar distances present in the distance matrix in descending order and summing the largest $n$ distances. Formally, let $D = \{d_{ij} \mid i, j \in V, i \neq j\}$ be the set of all unique travel distances between nodes, and let $D' = \langle d_{(1)}, d_{(2)}, \dots, d_{(|D|)} \rangle$ represent the sequence of these distances sorted in decreasing order (such that $d_{(1)} \geq d_{(2)} \geq \dots$). The upper bound $B$ is defined as:
\begin{equation}
B = \sum_{k=1}^{n} d_{(k)}
\label{eq:upper_bound}
\end{equation}
where $n$ represents the number of customers in the instance. This formulation guarantees a strict, robust upper limit that dynamically adapts to the spatial distribution of the specific benchmark instance.

A major computational advantage of both the Double Chromosome and Node Shift Encoding methods is that all stochastically generated offspring are structurally guaranteed to be feasible. This feasibility is maintained because genetic operators manipulate the underlying guiding vectors rather than operating directly on the physical customer sequence. 

Conversely, under the Path Representation, standard crossover and mutation operators frequently yield infeasible permutations (e.g., containing duplicate and omitted nodes), violating the visit uniqueness constraint. To strictly enforce feasibility within the population, a death penalty strategy is applied. Whenever a solution $s$ violates the Hamiltonian cycle constraints, its objective value $Z(s)$ is set equal to the upper bound $B$. This reduces its fitness value to exactly zero ($\text{Fitness}(s) = 0$), ensuring that such structurally invalid individuals are likely never selected to pass their genetic material to subsequent generations.

\subsubsection{Evolutionary operators}
To drive the evolutionary process, standard genetic operators are tailored to the respective representation schemes. First, a subset of surviving individuals is selected from the current population using the \textit{Roulette Wheel Selection} mechanism \cite{goldberg1989genetic}, which assigns selection probabilities proportional to each individual's fitness. To maintain population diversity and avoid premature convergence, the remainder of the intermediate population is generated completely at random. From this combined pool of survivors and randomized individuals, the final set of parents is selected using a \textit{Uniform Random Selection} to undergo evolutionary operations.

For the exploration of the search space, a standard \textit{One-Point Crossover} is applied to the selected parent pairs to generate offspring. In the case of the Path Representation, this operator is applied directly to the node sequences, whereas for the Double Chromosome and Node Shift Encoding (NSE) methods, the crossover operates strictly on the guiding or shift vectors, thereby preserving the structural validity of the decoded tours. 

Subsequently, genetic exploitation is balanced through a \textit{Simple Mutation} operator applied to a subset of the generated offspring, introducing stochastic perturbations to fine-tune the search. Finally, an elitism strategy is enforced by preserving the $p_1$ best individuals of the current population, which are reinserted into the next generation using tournament selection.

The step-by-step procedural framework of the implemented GA is outlined in Algorithm~\ref{GA_Implementation}.
		
\begin{algorithm}[ht!]
	\caption{Genetic Algorithm for Giant Tour Construction}
	\label{GA_Implementation} 
	\begin{algorithmic}[1]
	\State \textbf{Input:} Distance matrix
	\State \textbf{Output:} The optimized giant tour 			
	\State Initialize problem encoding using the selected representation (Path, Double Chromosome, or NSE)
	\State Generate a randomized initial population of size $N$
	\Repeat
	    \State Evaluate the fitness of each individual using Eq. (\ref{eq:fitness})		
	    \State Preserve $p_1$ elite individuals to seed the next generation
	    \State Select $p_2$ parents via Roulette Wheel Selection
	    \State Apply one-point crossover to the selected parents to generate offspring
	    \State Apply simple mutation to a randomly selected subset of $p_3$ offspring
	    \State Inject $p_4 = N-p_2$ randomly generated individuals to preserve population diversity
		\State Reinsert the $p_1$ elite individuals into the new population via tournament selection
	\Until{The maximum number of generations is reached}	
	\State \Return {The best giant tour individual found}
\end{algorithmic}
\end{algorithm}

\subsection{Tour splitting phase}
\label{subsect:splitting}
After the giant tour is constructed, the subsequent task is to partition  this global sequence into a set of feasible CVRP routes, each satisfying the vehicle capacity constraint $Q$. For this purpose, the linear splitting procedure developed by Vidal~\cite{Vidal2016} is utilized as the baseline partitioning mechanism.

The tour splitting phase aims to partition an uncapacitated giant tour $S = \langle s_1, s_2, \dots, s_n \rangle$ into a set of feasible routes. This partitioning problem was originally modeled by Beasley~\cite{Beasley1983} as a shortest path problem on an auxiliary Directed Acyclic Graph (DAG), denoted as $G = (V, A)$. Vidal~\cite{Vidal2016} later introduced an efficient method that solved this shortest path challenge in linear time $O(n)$.

In this network configuration, the depot is indexed by $0$, while the  $n$ customers of the giant tour are indexed from $1$ to $n$ in the order of their appearance. The node set of the DAG, $V = \{0, 1, \dots, n\}$, 
represents the potential cutting positions between customers. Here,  index $0$ represents the initial state at the depot, and node $i$ represents the state after visiting the $i$-th customer $s_i$ of the giant tour. An arc $(i, j) \in A$ exists if the sequence of consecutive customers $\langle s_{i+1}, \dots, s_j \rangle$ can be successfully served by a single vehicle without violating the maximum capacity constraint $Q$. The cost of an arc $(i,j)$, denoted as $w_{ij}$, represents the total distance of this isolated sub-tour, computed with:
\begin{equation}
w_{ij} = d_{0, s_{i+1}} + \sum_{k=i+1}^{j-1} d_{s_k, s_{k+1}} + d_{s_j, 0}
\end{equation}
where $d_{uv}$ represents the distance between nodes $u$ and $v$, and $0$ denotes the depot.

While the standard dynamic programming approach to find the shortest path in a DAG requires $O(n^2)$ operations, Vidal~\cite{Vidal2016} achieved a linear time complexity of $O(n)$ by exploiting the Monge property of the routing cost matrix. By maintaining a double-ended queue of candidate paths, dominated routes are systematically pruned. Let $V(j)$ be the minimum cumulative cost to partition the prefix $\langle s_1, \dots, s_j \rangle$ into feasible routes. The dynamic programming recurrence relation is defined as:
\begin{equation}
V(j) = \min_{i \in \{0, \dots, j-1\} \mid \text{feasible}(i,j)} \{ V(i) + w_{ij} \}
\label{eq:vidal_dp}
\end{equation}
subject to the boundary condition $V(0) = 0$. Following a single forward pass, the optimal total cost is obtained at $V(n)$, and the exact cluster boundaries are recovered via backtracking from node $n$ to node $0$.

Two partitioning strategies are considered in this work. The first employs the standard linear Split without fleet restriction, which may produce a partition with $|\mathcal{R}| > K$ routes. The second employs the $O(nK)$ extension of Vidal~\cite{Vidal2016} for limited fleets, which strictly enforces $|\mathcal{R}| \leq K$ but may fail to find any feasible 
partition when the constraints are jointly tight. In both cases, when the resulting number of routes exceeds the prescribed fleet size $K$, our specialized \textit{Machine Learning repair heuristic} is triggered as a corrective layer to restore feasibility. This heuristic is detailed in the subsequent section.

\subsection{Proposed Machine Learning repair heuristic}
\label{subsect:ml_repair_heuristic}

Hereafter, a Machine Learning (ML) post-processing correction heuristic is introduced to enhance the splitting phase and recover solution feasibility when the number of resultant routes exceeds the maximum number of available vehicles $K$ ($|\mathcal{R}| > K$). The repair mechanism operates as an unsupervised  online learning process: the route centroids act as adaptive prototypes that are sequentially updated after every customer insertion, and each assignment reshapes the proximity landscape for subsequent decisions. This dynamic adaptation, where the cluster model learns from prior assignments rather than relying on a fixed, precomputed rule, mirrors the update step of online k-means and other unsupervised  learning algorithms. In contrast to a static heuristic that would compute centroids once and assign customers permanently, our procedure refines its internal representation at every step, making it a genuine instance of learning-driven reallocation.

\subsubsection{Coordinate construction}
\label{coord_constr}
By using the CVRP benchmark distance matrix $D$, we construct a 2D Euclidean coordinate system. In fact, distance matrices derived from actual road layouts often deviate from flat 2D Euclidean metrics due to road curvature and terrain elevation changes. In our approach, we employ the \textit{distance geometry} embedding of Alencar et al.~\cite{Alencar2015} to derive the spatial coordinates $(x_i, y_i)$ of each node $i \in V$. This algorithm sequentially localizes nodes via sphere intersection (trilateration), solving a triangular linear system of order $d$ per iteration~\cite{Alencar2015}, where $d$ denotes the embedding dimension. For a fixed embedding dimension $d=2$, the localization cost per node is constant, leading to an overall linear complexity. 

\subsubsection{Centroid-based route reconstruction}
Once the 2D coordinates are extracted, the core strategy of the repair heuristic operates through a controlled destruction-reconstruction mechanism. It dynamically identifies the number of surplus routes, denoted as $m = |\mathcal{R}| - K$. To compress the fleet size without structural bias, the algorithm randomly selects exactly $m$ routes from the initial pool, dismantle them and move all their customers to an unassigned pool $\mathcal{U}$. The remaining $K$ active routes then serve as the destination clusters. 

To optimize the reconstruction phase, the heuristic framework emulates the behaviour of a $k$-means clustering mechanism, where spatial centroids ($C_r$) of the active routes act as dynamic cluster prototypes. It is worth noting that each centroid is computed solely from the customer coordinates assigned to that route, deliberately excluding the depot. The rationale is twofold. First, the depot is a fixed logistical origin shared by all routes; including it would artificially pull every centroid toward the origin, thereby collapsing the spatial diversity of cluster prototypes and depriving the reassignment rule of its discriminative power. Second, the centroid is meant to capture the geographical concentration of customers served by a route, so that proximity-based routing decisions remain meaningful. Were the depot included, the proximity measure would evaluate distance to a common hub rather than to the actual service zone, dispersing the search and rendering the spatial guidance ineffective. The proximity to a cluster centre guides the insertion, and the centroid coordinates are immediately recalculated upon each customer integration to maintain topographical accuracy. We refer to this approach as \textit{MLH}. The sequential operational steps are formally detailed in Algorithm~\ref{ML_Heuristic_algo}.

\begin{algorithm}[ht!]
\caption{MLH for route reassignment}
\label{ML_Heuristic_algo}
\begin{algorithmic}[1]	
	\State \textbf{Input:} Initial routes $\mathcal{R}$, distance matrix $D$, customer demands $\{q_i\}$, maximum allowed vehicles $K$, vehicle capacity $Q$
	\State \textbf{Output:} Optimized set of physical routes
	\State Extract geospatial coordinates $(x_i, y_i)$ for all clients $i = 2, \dots, n$ using Alencar's embedding \cite{Alencar2015}
	\State $\mathcal{R}_{\text{old}} \gets \mathcal{R}$ 
	\State $m \gets |\mathcal{R}| - K$ 
	\State $\mathcal{U} \gets \emptyset$ 
	
	\For{$j = 1$ \textbf{to} $m$}
	    \State Select a route $r \in \mathcal{R}$ randomly
	    \State $\mathcal{U} \gets \mathcal{U} \cup \{i \mid i \in r\}$ 
	    \State $\mathcal{R} \gets \mathcal{R} \setminus \{r\}$ 
	\EndFor
	
	\For{\textbf{each} route $r \in \mathcal{R}$}
		\State Compute the spatial centroid $C_r$ using $r$'s customer coordinates $(x_i, y_i)$
	\EndFor
	
	\State $\mathcal{U}_{\text{sorted}} \gets$ Sort $\mathcal{U}$ in descending order of customer demand $q_i$
	
	\For{\textbf{each} client $i$ in $\mathcal{U}_{\text{sorted}}$}
		\State $\mathcal{S} \gets \{ r \in \mathcal{R} \mid Q_r + q_i \le Q \}$ 
		\If{$\mathcal{S} \neq \emptyset$}
	        \State $\mathcal{S}_{\text{sorted}} \gets$ Sort $\mathcal{S}$ in ascending order of distance to centroid $C_r$
		    \State $r^* \gets$ First element of $\mathcal{S}_{\text{sorted}}$
		    \State Insert customer $i$ into $r^*$ at the position that minimizes the travelled distance
		    \State $Q_{r^*} \gets Q_{r^*} + q_i$ 
            \State Recalculate spatial centroid $C_{r^*}$ of route $r^*$
	   \Else
	        \State \textbf{return} $\mathcal{R}_{\text{old}}$ 
	   \EndIf
	\EndFor	
	\State \textbf{return} $\mathcal{R}$		
\end{algorithmic}
\end{algorithm}	

\subsection{Complexity analysis}

The repair heuristic (algorithm~\ref{ML_Heuristic_algo}) spends its time in three distinct phases.
\begin{itemize}
\item Spatial embedding and pool construction (lines~3--14)
Extracting the 2D coordinates via Alencar et al.'s embedding requires $O(n)$ time for fixed dimension $d=2$, since each node is localised by solving a triangular linear system of order two (section~\ref{coord_constr}). Dismantling the $m=|\mathcal{R}|-K$ surplus routes and building the unassigned pool $\mathcal{U}$ touches each displaced customer once, which costs $O(|\mathcal{U}|)$. Computing the initial spatial centroids of the $K$ surviving routes then requires a single pass over all customers that remain assigned, adding another $O(n)$. 

\item Sorting $\mathcal{U}$ (line~15) by decreasing demand is a standard comparison sort and therefore costs $O(|\mathcal{U}|\log|\mathcal{U}|)$.

\item Reassignment loop (lines~16--27) 
For each client $i\in\mathcal{U}_{\text{sorted}}$, capacity filtering scans at most $K$ routes. Sorting the feasible candidates by their distance to the route centroids adds $O(K\log K)$. Locating the best insertion position in the selected route $r^{*}$ requires examining the sequence of that route, which on average contains $n/K$ customers. The centroid of $r^{*}$ is updated incrementally after each insertion, so its recalculation is $O(1)$. Across all $|\mathcal{U}|$ clients, the loop therefore contributes $O(|\mathcal{U}|\cdot(K\log K + n/K))$.

\item Overall complexity
Combining the three phases yields:
\begin{equation}
O\!\left(n + |\mathcal{U}|\log|\mathcal{U}| + |\mathcal{U}|\cdot\Big(K\log K + \frac{n}{K}\Big)\right)
\label{eq:mlh-complexity}
\end{equation}

The worst theoretical case is characterized by the presence of a very large number of clients to be served by very few high-capacity vehicles. In this extreme scenario, the split procedure might produce an arbitrarily large number of surplus routes, causing $|\mathcal{U}|$ to reach ${O}(n)$ and degrading the overall complexity to $O(n^2)$. However, such an extreme scenario is highly unlikely in practice. Instead, the unconstrained linear split partitions the giant tour into capacity-feasible segments, each strictly respecting the vehicle capacity limit $Q$; thus, no single route can contain an arbitrary number of customers. Because the route surplus $m = |\mathcal{R}| - K$ remains bounded by a small constant across all realistic instances, $|\mathcal{U}|$ stays modest, ensuring that the heuristic operates as a lightweight post-processing layer whose runtime is dominated by the linear embedding term.
\end{itemize}
\subsubsection{Illustrative example}
To make the mechanics of the ML repair heuristic concrete, we walk through a small instance with thirteen nodes: one depot and twelve customers labelled A to L. The distance matrix is given below:
{\scriptsize
\[
\mathbf{D} = 
\left[
\begin{array}{ccccccccccccc}
0 & 120 & 160 & 200 & 50 & 150 & 100 & 124 & 100 & 100 & 171 & 80 & 144  \\
120 &  0 & 200 & 160 & 100 & 124 & 50 & 150 & 100 & 60 & 171 & 100 & 80  \\
160 & 200 &  0 & 120 & 124 & 50 & 150 & 50 & 100 & 171 & 60 & 80 & 144  \\ 
200 & 160 & 120 &  0 & 150 & 50 & 124 & 100 & 100 & 171 & 60 & 144 & 80  \\ 
50 & 100 & 124 & 150 &  0 & 100 & 60 & 80 & 50 & 50 & 124 & 50 & 100  \\ 
150 & 124 & 50 & 50 & 100 &  0 & 80 & 60 & 50 & 124 & 50 & 100 & 50  \\ 
100 & 50 & 150 & 124 & 60 & 80 &  0 & 100 & 50 & 50 & 124 & 100 & 50  \\ 
124 & 150 & 50 & 100 & 80 & 60 & 100 &  0 & 50 & 124 & 50 & 50 & 100  \\ 
100 & 100 & 100 & 100 & 50 & 50 & 50 & 50 &  0 & 80 & 80 & 60 & 60  \\ 
100 & 60 & 171 & 171 & 50 & 124 & 50 & 124 & 80 &  0 & 160 & 100 & 100  \\
171 & 171 & 60 & 60 & 124 & 50 & 124 & 50 & 80 & 160 &  0 & 100 & 100  \\ 
80 & 100 & 80 & 144 & 50 & 100 & 100 & 50 & 60 & 100 & 100 &  0 & 120  \\ 
144 & 80 & 144 & 80 & 100 & 50 & 50 & 100 & 60 & 100 & 100 & 120 &  0  

\end{array}
\right]
\]
}
Vehicle capacity is Q=15  and the fleet is limited to K=3  vehicles. Customer demands appear in Table~\ref{tab:customers_demands}.

\begin{table}[ht!]
\caption{Customer demands}
\label{tab:customers_demands}
\begin{tabular*}{\linewidth}{@{\extracolsep{\fill}}lccccccccccccc@{}}
\toprule
Node & Depot & A & B & C & D & E & F & G & H & I & J & K & L \\
\midrule
Demand & 0 & 5 & 4 & 4 & 3 & 6 & 2 & 5 & 5 & 1 & 4 & 3 & 2 \\
\bottomrule
\end{tabular*}
\end{table}

\begin{figure}[ht!]
  \centering
  \includegraphics[width=0.4\linewidth]{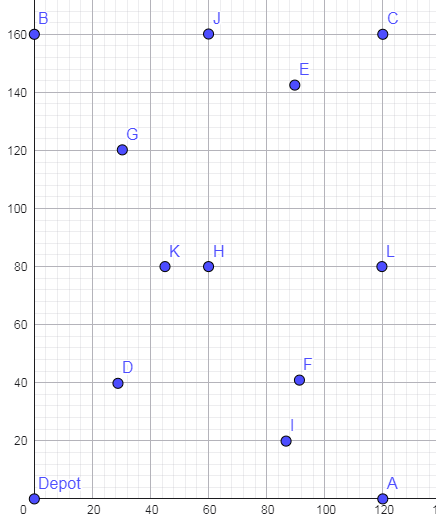}
  \caption{Illustrative example 2D map customer localisation}
  \label{illustExample}
\end{figure}

First of all, we derive the 2D coordinates of each node using the distance-geometry embedding. The resulting coordinates are shown in Table~\ref{tab:customer_coordinates} and the corresponding 2D map is depicted in Figure \ref{illustExample}.

\begin{table}[ht!]
\caption{Reconstructed Cartesian coordinates of nodes}
\label{tab:customer_coordinates}
\begin{tabular*}{0.5\textwidth}{@{}lll@{}}
\toprule
Node & X & Y \\
\midrule
Depot                   & 0.0          & 0.0          \\
A                       & 120.0        & 0.0          \\
B                       & 0.0          & 160.0        \\
C                       & 120.0        & 160.0        \\
D                       & 28.8         & 39.8         \\
E                       & 89.7         & 142.5        \\
F                       & 91.3         & 40.9         \\
G                       & 30.3         & 120.2        \\
H                       & 60.0         & 80.0         \\
I                       & 86.7         & 19.9         \\
J                       & 60.0         & 160.1        \\
K                       & 45.0         & 80.0         \\
L                       & 119.7        & 80.0         \\
\bottomrule
\end{tabular*}
\end{table}

For instance, if the initial solution produced by the splitting phase consists of four routes, exceeding the fleet limit $K = 3$, as follows:
\begin{align*}
R_1 &= \{Depot, K(3), H(5), D(3), Depot\},   & Q_1 &= 11, \\
R_2 &= \{Depot,J(4), E(6), C(4) , Depot\},   & Q_2 &= 14, \\
R_3 &= \{Depot,I(1), A(5), F(2), L(2), Depot\},   & Q_3 &= 10, \\
R_4 &= \{Depot,B(4), G(5), Depot\},   & Q_4 &= 9, \\
\end{align*}
All routes satisfy the capacity constraint, but $|\mathcal{R}| = 5 > K = 3$. The surplus is $m = |\mathcal{R}| - K = 1$ route. For example, route $R_2$ is randomly selected and dismantled. Its customers are transferred to the unassigned pool $\mathcal{U}$ and sorted in descending order of demand:
\[
\mathcal{U}_{\text{sorted}} = \{E(6), J(4), C(4)\}.
\]

\begin{figure}[ht!]
  \centering
  \includegraphics[width=0.4\linewidth]{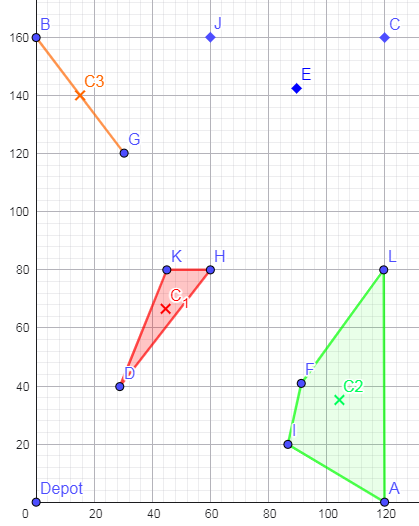}
  \caption{Pre-assignment phase}
  \label{illustExampleCentroids}
\end{figure}

The remaining routes $R_1, R_3, R_4$ serve as the destination clusters ($R_1, R_2, R_3$). To guide reassignment, each cluster is endowed with a spatial centroid derived from two-dimensional coordinates.
The initial spatial centroids of the three sets of customers associated to the active routes are therefore:
\[
C_1 = (44.60, 66.60), \quad C_2 = (104.43, 35.20), \quad C_3 = (15.15, 140.10).
\]
as depicted in Figure \ref{illustExampleCentroids}.

For each customer in $\mathcal{U}_{\text{sorted}}$, the Euclidean distance to the current centroids of the active routes is computed. The customer is inserted into the nearest route that preserves the capacity constraint; the centroid of the selected route is then immediately updated to establish the topographical change for the next assignment. Table~\ref{tab:reassignment} details the sequential reassignment.

\begin{table}[ht!]
\caption{Sequential centroid-guided reassignment with dynamic centroid updates ($Q = 15$).}
\label{tab:reassignment}
\begin{tabular*}{\linewidth}{@{\extracolsep{\fill}}lllllllll@{}}
\toprule
Cust. & Dem. & $d(C_1)$ & $d(C_2)$ & $d(C_3)$ & Priority & Decision & Load & New Centroid \\
\midrule
$E$ & 6 & 88.29 & 108.31 & \textbf{74.59} & $R_3, R_1, R_2$ & $E \to R_3$ & $R_3 = 15$ & $C_3 = (40.00, 140.90)$ \\
$J$ & 4 & \textbf{94.76} & 132.57 & -- & $R_1, R_2$ & $J \to R_1$ & $R_1 = 15$ & $C_1 = (48.45, 89.98)$ \\
$C$ & 4 & -- & \textbf{125.77} & -- & $R_2$ & $C \to R_2$ & $R_2 = 14$ & $C_2 = (107.54, 60.16)$ \\
\bottomrule
\end{tabular*}
\end{table}

\begin{figure}[ht!]
  \centering
  \includegraphics[width=0.4\linewidth]{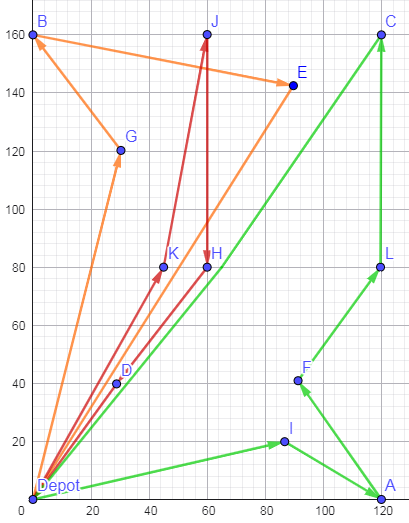}
  \caption{Final routes after reassignment}
  \label{fig:final_routes}
\end{figure}

The final solution compresses the fleet to exactly $K = 3$ routes, all capacity-feasible. Each reassigned customer is inserted at the position within its designated route that minimizes the incremental travel distance (see Fig.\ref{fig:final_routes}):
\begin{align*}
R_1 &: Depot \to K \to j \to H \to D \to Depot,  & Q_1 &= 15 \leq 15, \\
R_2 &: Depot \to I \to A \to F \to L \to C \to Depot,  & Q_2 &= 14 \leq 15, \\
R_3 &: Depot \to G \to B \to E \to Depot,  & Q_3 &= 15 \leq 15.
\end{align*}

\section{Experimental study}
\label{sect:Exper}

This section evaluates and compares the computational performance of the three investigated encoding schemes and the proposed ML-heuristic. We first describe the benchmark configurations, followed by the parameter tuning protocol and the penalty framework established to handle fleet capacity violations.

\subsection{Benchmark instances}
To evaluate the performances of the three alternative encoding schemes, a standard benchmark suite proposed by Augerat et al. \cite{Augerat1995} was employed. This dataset is widely recognized in the CVRP literature as it features a diverse mix of different problem scales, customer distributions, and capacity constraint tightness. For each instance, the exact number of customers, the physical vehicle availability constraints, and the homogeneous vehicle capacities are detailed in Table~\ref{tab:augerat-instances}.

\begin{table}[ht!]
\caption{Summary of the benchmark instances}
\label{tab:augerat-instances}
\begin{tabular*}{\linewidth}{@{\extracolsep{\fill}}clccc@{}}
\toprule
\textbf{No.} & \textbf{Instance} & \textbf{Customers} & \textbf{Max Vehicles ($K$)} & \textbf{Vehicle Capacity ($Q$)} \\
\midrule
 1  & P-n16-k8   & 15  & 8   & 35    \\
 2  & P-n19-k2   & 18  & 2   & 160   \\
 3  & P-n20-k2   & 19  & 2   & 160   \\
 4  & P-n21-k2   & 20  & 2   & 160   \\
 5  & P-n22-k2   & 21  & 2   & 160   \\
 6  & P-n22-k8   & 21  & 8   & 3000  \\
 7  & P-n23-k8   & 22  & 8   & 40    \\
 8  & P-n40-k5   & 39  & 5   & 140   \\
 9  & P-n45-k5   & 44  & 5   & 150   \\
 10 & P-n50-k7   & 49  & 7   & 150   \\
 11 & P-n50-k8   & 49  & 8   & 120   \\
 12 & P-n50-k10  & 49  & 10  & 100   \\
 13 & P-n51-k10  & 50  & 10  & 80    \\
 14 & P-n55-k7   & 54  & 7   & 170   \\
 15 & P-n55-k8   & 54  & 8   & 160   \\
 16 & P-n55-k10  & 54  & 10  & 115   \\
 17 & P-n55-k15  & 54  & 15  & 70    \\
 18 & P-n60-k10  & 59  & 10  & 120   \\
 19 & P-n60-k15  & 59  & 15  & 80    \\
 20 & P-n65-k10  & 64  & 10  & 130   \\
 21 & P-n70-k10  & 69  & 10  & 135   \\
 22 & P-n76-k4   & 75  & 4   & 350   \\
 23 & P-n76-k5   & 75  & 5   & 280   \\
 24 & P-n101-k4  & 100 & 4   & 400  \\
\bottomrule
\end{tabular*}
\end{table}

\subsection{Experimental protocol}
\label{exp_protocol}
To determine the optimal configuration for the execution layers, comprehensive experimental combinations were tested for each encoding scheme. The operational parameters selected for the comparative analysis correspond to those exhibiting the best overall trade-off between final solution convergence speed and absolute solution quality.

To secure a statistically valid evaluation, each representation model—namely Path Representation, Double Chromosome, and Node Shift Encoding—was executed independently 30 times per benchmark instance, mitigating the stochastic variation inherent to Genetic Algorithms.

The individual runs yielded a distribution of both feasible and infeasible routing structures. Infraction of solution feasibility typically arose when the total number of physical routes obtained during the splitting phase strictly exceeded the vehicle availability fleet size ($K$) specified by the instance data. To address this structural issue, the Machine Learning post-processing repair heuristic detailed in Section~\ref{subsect:ml_repair_heuristic} was triggered to restore solution compliance by reallocating customers from randomly selected $m$ routes into active clusters based on proximity and capacity.  When spatial containment constraints make repair impossible, an experimentally derived upper bound was employed to penalize the solution.
			
\subsection{Implementation environment}
The metaheuristic framework was developed and compiled in C\# within the .NET framework environment. All experimental evaluations were executed on a dedicated Intel Core i3 processor operating at 2.00 GHz with 4 GB of RAM. The numerical findings obtained under these controlled operating conditions are presented in the section that follows.

\section{Results}
\label{sect:Res}

This section presents the computational results obtained from the experimental evaluation. We first analyse the performance of the three encoding strategies (NSE, DC, PR) before applying the ML repair heuristic, then examine its impact on feasibility restoration. Finally, we compare \textit{Linear Split + ML heuristic} pipeline against \textit{Split with Limited Fleet} strategy and our exact mathematical formulation (see Section \ref{sect:PM}) solved with a time limit using the GNU Linear Programming Kit (GLPK) \cite{glpk}.

\subsection{Encoding comparison}
\label{Enc_comp}		
Table~\ref{tab:results} reports the comparative performance of the NSE, DC, and PR encoding strategies across 24 benchmark instances. For each strategy, the following metrics are provided: best solution cost (Best), average solution cost (Average), worst solution cost (Worst), average computational time in milliseconds (Time), the number of feasible solutions obtained out of 30 runs (FC), and the average fleet size violation ($K_{err}$), defined as the mean number of vehicles exceeding the prescribed maximum $K$.

\begin{table}[ht!]
\centering
\footnotesize
\setlength{\tabcolsep}{0.9pt}
\caption{Results of the encoding strategies on the benchmark instances}
\label{tab:results}

\begin{tabular}{ll llllll llllll llllll}
\toprule
\multirow{2}{*}{\textbf{ID}} & \multirow{2}{*}{\textbf{Instance}} 
& \multicolumn{6}{c}{\textbf{NSE}} 
& \multicolumn{6}{c}{\textbf{DC}} 
& \multicolumn{6}{c}{\textbf{PR}} 
\\
\cmidrule(lr){3-8} \cmidrule(lr){9-14} \cmidrule(lr){15-20}
& & {Best} & {Average} & {Worst} & {Time (ms)} & {FC} & {$K_{err}$} 
  & {Best} & {Average} & {Worst} & {Time (ms)} & {FC} & {$K_{err}$} 
  & {Best} & {Average} & {Worst} & {Time (ms)} & {FC} & {$K_{err}$} \\
\midrule
1  & P-n16-k8     & \textbf{456.95} & 842.94 & 856.25 & 677   & 1  & \textbf{1}    & 856.25 & 856.25 & 856.25 & 436   & 0 & 1.33 & 464.39 & 831.33 & 856.25 & 717   & \textbf{2} & 1.43 \\
2  & P-n19-k2     & \textbf{244.60} & 662.19 & 833.66 & 689   & \textbf{9}  & \textbf{0.7}  & 276.44 & 692.45 & 833.66 & 364   & 8 & 0.73 & 319.63 & 703.04 & 833.66 & 681   & 8 & 0.73 \\
3  & P-n20-k2     & \textbf{244.69} & 510.33 & 870.56 & 467   & \textbf{18} & \textbf{0.4}  & 270.23 & 592.45 & 870.56 & 398   & 15& 0.5  & 330.41 & 702.67 & 870.56 & 641   & 10& 0.66  \\
4  & P-n21-k2     & \textbf{262.44} & 455.02 & 907.35 & 470   & \textbf{22} & \textbf{0.26} & 276.15 & 555.77 & 907.35 & 374   & 18& 0.4  & 351.88 & 593.20 & 907.35 & 345   & 18& 0.4   \\
5  & P-n22-k2     & \textbf{234.92} & 593.72 & 950.22 & 685   & \textbf{16} & \textbf{0.46} & 284.80 & 699.12 & 950.22 & 382   & 12& 0.6  & 380.77 & 709.70 & 950.22 & 342   & 13& 0.56 \\
6  & P-n22-k8     & \textbf{2007.05} & 2007.05 & 2007.05 & 907   & \textbf{0}  & 2.26 & \textbf{2007.05} & 2007.05 & 2007.05 & 379   & \textbf{0} & \textbf{2.16} & \textbf{2007.05} & 2007.05 & 2007.05 & 342   & \textbf{0} & 2.2   \\
7  & P-n23-k8     & \textbf{1212.67} & 1212.67 & 1212.67 & 529   & \textbf{0}  & \textbf{2.43} & \textbf{1212.67} & 1212.67 & 1212.67 & 375   & \textbf{0} & 2.80 & \textbf{1212.67} & 1212.67 & 1212.67 & 345   & \textbf{0} & 2.60 \\
8  & P-n40-k5     & \textbf{669.61} & 2272.03 & 2647.84 & 3652  & \textbf{6}  & 0.96 & 788.68 & 2288.56 & 2647.84 & 1923  & \textbf{6} & \textbf{0.86} & 994.28 & 2539.26 & 2647.84 & 2379  & 2 & 1.23 \\
9  & P-n45-k5     & \textbf{1036.76} & 3158.56 & 3231.73 & 3863  & \textbf{1}  & \textbf{1.33} & 3231.73 & 3231.73 & 3231.73 & 1990  & 0 & 1.43 & 3231.73 & 3231.73 & 3231.73 & 3015  & 0 & 1.63 \\
10 & P-n50-k7     & 3421.81 & 3421.81 & 3421.81 & 4194  & 0  & \textbf{1.5}  & \textbf{1051.46} & 3342.8 & 3421.81 & 2137  & \textbf{1} & 1.6  & 3421.81 & 3421.81 & 3421.81 & 2505  & 0 & 1.7  \\
11 & P-n50-k8     & \textbf{3478.25} & 3478.25 & 3478.25 & 4061  & \textbf{0}  & \textbf{2.56} & \textbf{3478.25} & 3478.25 & 3478.25 & 1911  & \textbf{0} & 2.66 & \textbf{3478.25} & 3478.25 & 3478.25 & 2136  & \textbf{0} & 2.93 \\
12 & P-n50-k10    & \textbf{3591.08} & 3591.08 & 3591.08 & 4379  & \textbf{0}  & \textbf{2.66} & \textbf{3591.08} & 3591.08 & 3591.08 & 1954  & \textbf{0} & 2.93 & \textbf{3591.08} & 3591.08 & 3591.08 & 2189  & \textbf{0} & 2.9  \\
13 & P-n51-k10    & \textbf{3894.33} & 3894.33 & 3894.33 & 4303  & \textbf{0}  & \textbf{2.63} & \textbf{3894.33} & 3894.33 & 3894.33 & 2066  & \textbf{0} & 2.96 & \textbf{3894.33} & 3894.33 & 3894.33 & 2205  & \textbf{0} & 2.93 \\
14 & P-n55-k7     & \textbf{1049.09} & 3562.65 & 3738.75 & 4606  & \textbf{2}  & 1.5  & 1142.60 & 3566.02 & 3738.75 & 2105  & \textbf{2} & \textbf{1.4}  & 1400.57 & 3660.81 & 3738.75 & 2460  & 1 & 1.5 \\
15 & P-n55-k8     & \textbf{947.10} & 2788.01 & 3795.76 & 5032  & \textbf{11} & \textbf{0.76} & 1036.14 & 2292.87 & 3795.76 & 1951  & 9 & 0.8  & 1366.51 & 3316.70 & 3795.76 & 2691  & 6 & 1.26 \\
16 & P-n55-k10    & \textbf{3909.47} & 3909.47 & 3909.47 & 4637  & \textbf{0}  & \textbf{2.06} & \textbf{3909.47} & 3909.47 & 3909.47 & 2105  & \textbf{0} & \textbf{2.06} & \textbf{3909.47} & 3909.47 & 3909.47 & 2209  & \textbf{0} & 2.3  \\
17 & P-n55-k15    & \textbf{4192.28} & 4192.28 & 4192.28 & 4678  & \textbf{0}  & 4.66 & \textbf{4192.28} & 4192.28 & 4192.28 & 1854  & \textbf{0} & \textbf{4.53} & \textbf{4192.28} & 4192.28 & 4192.28 & 2226  & \textbf{0} & 5.2  \\
18 & P-n60-k10    & \textbf{4430.10} & 4430.10 & 4430.10 & 11626 & \textbf{0}  & \textbf{2.36} & \textbf{4430.10} & 4430.10 & 4430.10 & 6424  & \textbf{0} & \textbf{2.36} & \textbf{4430.10} & 4430.10 & 4430.10 & 6557  & \textbf{0} & 2.9 \\
19 & P-n60-k15    & \textbf{4758.57} & 4758.57 & 4758.57 & 10991 & \textbf{0}  & \textbf{3.53} & \textbf{4758.57} & 4758.57 & 4758.57 & 6213  & \textbf{0} & 3.86 & \textbf{4758.57} & 4758.57 & 4758.57 & 6288  & \textbf{0} & 3.66 \\
20 & P-n65-k10    & \textbf{5002.05} & 5002.05 & 5002.05 & 11617 & \textbf{0}  & 2.53 & \textbf{5002.05} & 5002.05 & 5002.05 & 6696  & \textbf{0} & \textbf{2.5}  & \textbf{5002.05} & 5002.05 & 5002.05 & 6275  & \textbf{0} & 3 \\
21 & P-n70-k10    & \textbf{5449.88} & 5449.88 & 5449.88 & 12984 & \textbf{0}  & 3    & \textbf{5449.88} & 5449.88 & 5449.88 & 6774  & \textbf{0} & \textbf{2.93} & \textbf{5449.88} & 5449.88 & 5449.88 & 6405  & \textbf{0} & 3.16 \\
22 & P-n76-k4     & 5566.67 & 5566.67 & 5566.67 & 47849 & 0  & 1.46 & 5566.67 & 5566.67 & 5566.67 & 30126 & 0 & 1.7  & \textbf{2205.91} & 5454.65 & 5566.67 & 25363 & \textbf{1} & \textbf{1.43} \\
23 & P-n76-k5     & \textbf{5631.30} & 5631.30 & 5631.30 & 54207 & \textbf{0}  & \textbf{1.76} & \textbf{5631.30} & 5631.30 & 5631.30 & 31396 & \textbf{0} & \textbf{1.76} & \textbf{5631.30} & 5631.30 & 5631.30 & 25477 & \textbf{0} & 1.80 \\
24 & P-n101-k4    & \textbf{1681.09} & 6874.50 & 7655.51 & 66420 & \textbf{4}  & 1.13 & 2123.19 & 7471.10 & 7655.51 & 33916 & 1 & 1.16 & 3030.55 & 7045.86 & 7655.51 & 28129 & \textbf{4} & \textbf{1.10} \\
\bottomrule
\end{tabular}

\end{table}

The results reveal a clear hierarchy among the three encoding strategies when coupled with the linear Split algorithm. The analysis proceeds along two dimensions: solution quality and feasibility performance.

\begin{figure}[ht!]
\centering
\includegraphics[width=\textwidth]{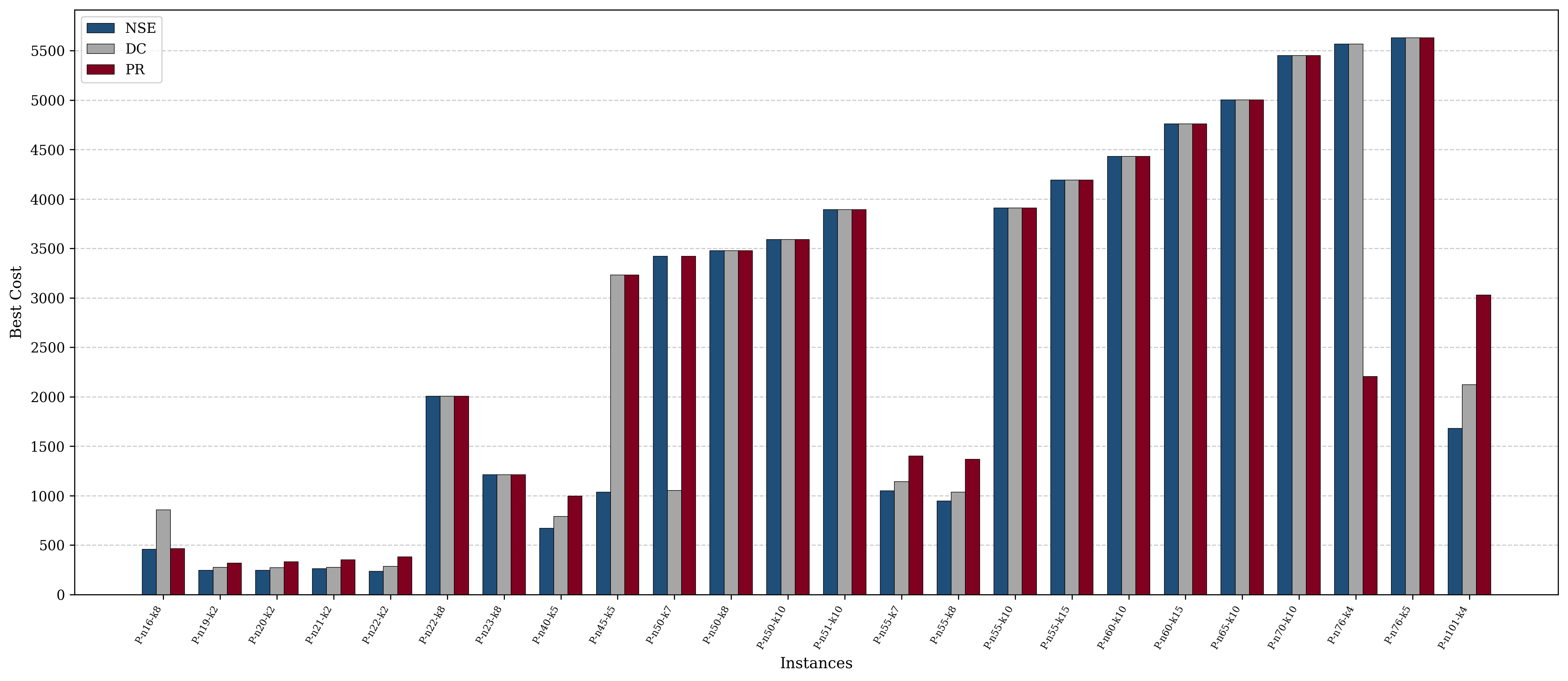}
\caption{Best solution cost comparison across instances}
\label{fig:fitness_comp}
\end{figure}

Regarding solution quality(Fig. \ref{fig:fitness_comp}), NSE consistently dominates across all instances, whether feasible or infeasible. On the 24 benchmark instances, NSE achieves the strictly lowest Best cost on 10 instances, compared to only 1 for DC and 1 for PR, with the remaining 12 instances being ties. This superiority is particularly pronounced on P-n20-k2 where NSE reaches 244.69 against 270.23 for DC and 330.41 for PR, on P-n22-k2 with 234.92 against 284.80 and 380.77 respectively, and on P-n101-k4 with 1681.09 against 2123.19 and 3030.55. 

\begin{figure}[ht!]
\centering
\includegraphics[width=\textwidth]{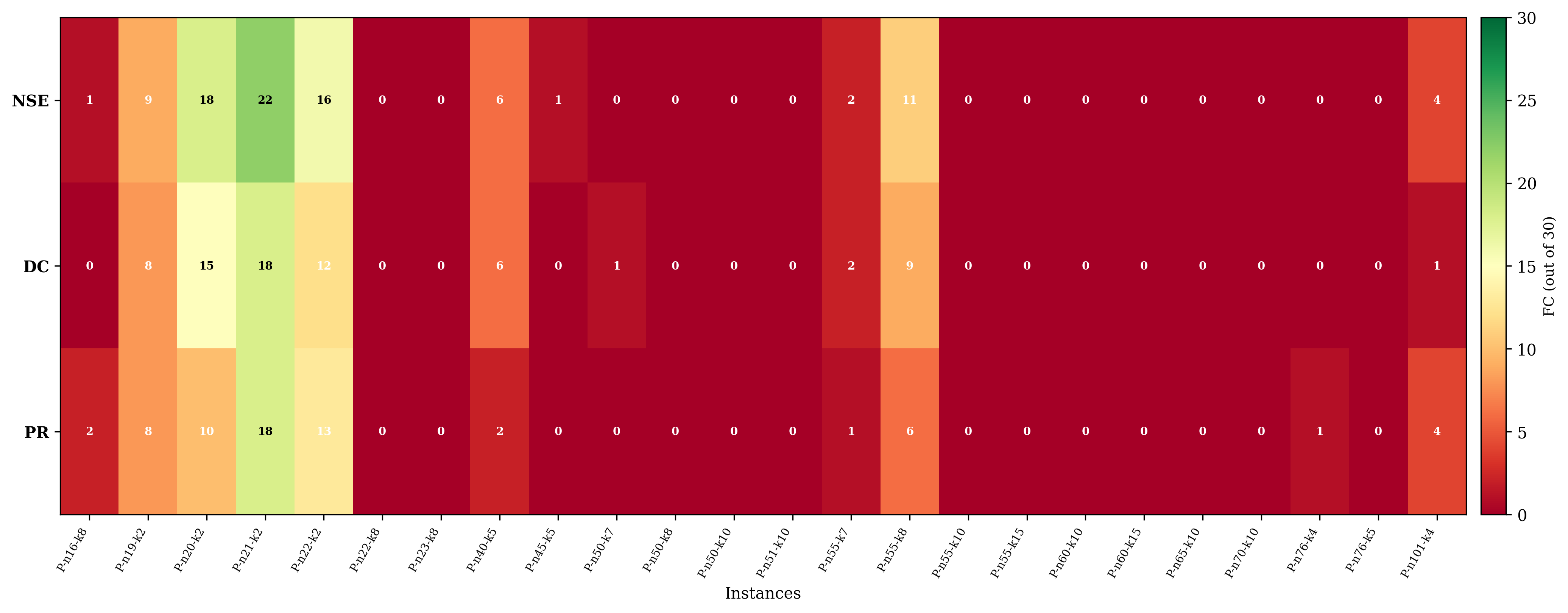}
\caption{Feasibility rate heatmap for NSE, DC, and PR encoding strategies}
\label{fig:feasibility_heatmap}
\end{figure}

The feasibility analysis reinforces this dominance. NSE achieves the highest number of feasible solutions (FC) on 6 instances, compared to 2 for DC and 2 for PR and one tie with PR and two with DC. On P-n20-k2, NSE delivers 18 feasible solutions out of 30 runs against 15 for DC and 10 for PR; on P-n21-k2, NSE reaches 22 against 18 for both competitors; on P-n22-k2, NSE achieves 16 against 12 for DC and 13 for PR (Fig. \ref{fig:feasibility_heatmap}). 

\begin{figure}[ht!]
\centering
\includegraphics[width=\textwidth]{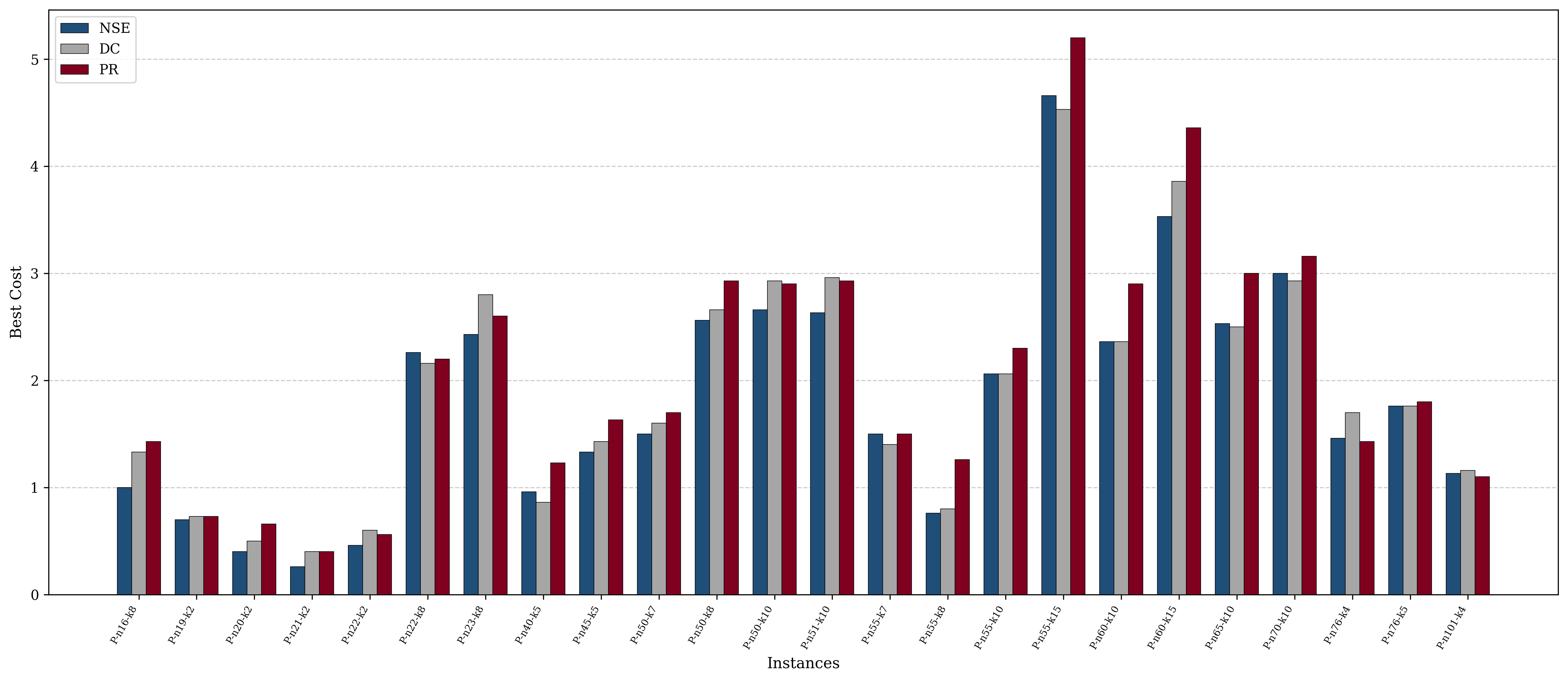}
\caption{Average vehicle excess for each instance}
\label{fig:k_err}
\end{figure}

More importantly, the $K_{err}$ metric, which measures the average number of vehicles exceeding the fleet limit,is systematically lowest for NSE. Across the 24 instances, NSE achieves the strictly lowest $K_{err}$ on 13 instances, compared to 6 for DC and 1 for PR. NSE ties with DC on 3 instances (P-n55-k10, P-n60-k10, and P-n76-k5) and ties with PR on 1 instance (P-n76-k5). This is critical because a lower $K_{err}$ means that the solution has fewer excess routes to eliminate, making the correction process both faster and more likely to succeed. On instances where no feasible solution is found, NSE's $K_{err}$ remains controlled: on P-n50-k8, NSE shows 2.56 against 2.66 for DC and 2.93 for PR; on P-n60-k15, NSE records 3.53 against 3.86 for DC and 4.36 for PR(see Fig.\ref{fig:k_err}).

\begin{figure}[ht!]
\centering
\includegraphics[width=\textwidth]{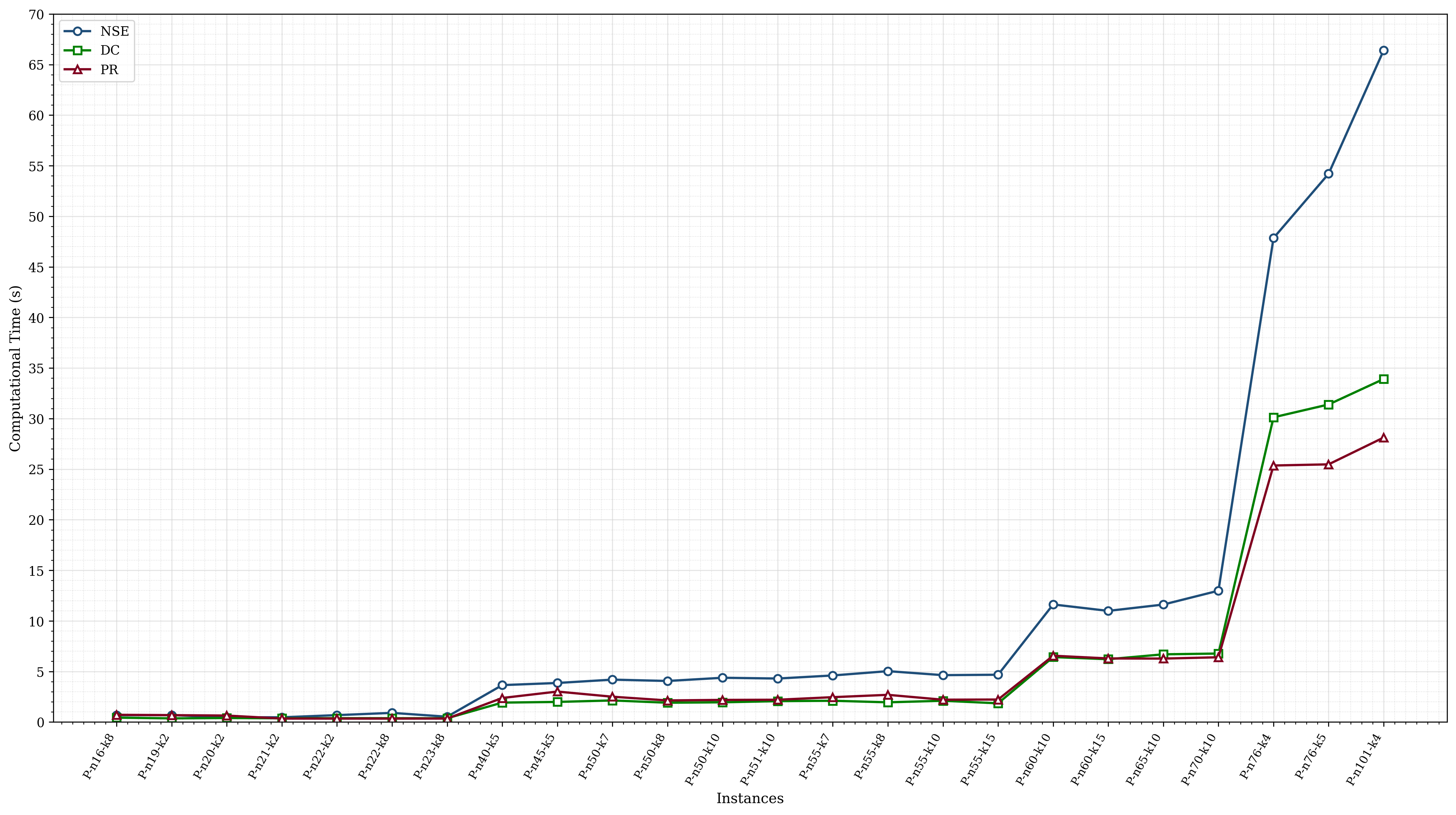}
\caption{Runtime scaling of the three encoding strategies.}
\label{fig:runtime_curves}
\end{figure}

The computational time reveals a different ranking, with DC being the fastest encoding on 15 instances, while PR is fastest on the remaining 9 instances. However, the time gap is not prohibitive for NSE: on P-n101-k4, NSE requires 66,420 ms against 33,916 ms for DC and 28,129 ms for PR (see Fig. \ref{fig:runtime_curves}). 

The synthesis of these results is unambiguous. NSE emerges as the superior encoding strategy across both feasible and infeasible regimes. It produces the best and average solution costs, the highest feasibility rates, and critically, the lowest fleet violations when constraints are not satisfied. 

\subsection{ML heuristic effect} 

To address the fleet infeasibility observed in Table~\ref{tab:results}, the ML repair heuristic is applied to the three encoding strategies. Table~\ref{tab:results_after_MLH} reports the resulting performance, where each encoding is now coupled with \textit{Linear Split} followed by the proposed heuristic whenever infeasibility occurs.

\begin{table}[ht!]
\centering
\footnotesize
\setlength{\tabcolsep}{0.9pt}
\caption{Results of the encoding strategies with the ML-heuristic}
\label{tab:results_after_MLH}

\begin{tabular}{ll llllll llllll llllll}
\toprule
\multirow{2}{*}{\textbf{ID}} & \multirow{2}{*}{\textbf{Instance}} 
& \multicolumn{6}{c}{\textbf{NSE}} 
& \multicolumn{6}{c}{\textbf{DC}} 
& \multicolumn{6}{c}{\textbf{PR}} 
\\
\cmidrule(lr){3-8} \cmidrule(lr){9-14} \cmidrule(lr){15-20}
& & {Best} & {Average} & {Worst} & {Time (ms)} & {FC} & {$K_{err}$} 
  & {Best} & {Average} & {Worst} & {Time (ms)} & {FC} & {$K_{err}$} 
  & {Best} & {Average} & {Worst} & {Time (ms)} & {FC} & {$K_{err}$} \\
\midrule
1 & P-n16-k8 & \textbf{456.95} & 653.5 & 856.25 & 677.08 & \textbf{16} & \textbf{0.47} & 470.78 & 685.46 & 856.25 & 436.07 & 14 & 0.57 & 464.39 & 750.2 & 856.25 & 717.08 & 9 & 0.83 \\
2 & P-n19-k2 & 244.6 & 327.64 & 833.66 & 689.1 & 27 & 0.1 & \textbf{239.76} & \textbf{315.59} & 833.66 & 364.06 & \textbf{29} & \textbf{0.03} & 263.99 & 355.17 & 833.66 & 681.08 & 28 & 0.07 \\
3 & P-n20-k2 & \textbf{244.69} & 312.97 & 870.56 & 467.05 & 28 & 0.07 & 252.09 & 341.27 & 870.56 & 398.04 & 28 & 0.07 & 268.9 & 358.46 & 870.56 & 641.08 & \textbf{29} & \textbf{0.03} \\
4 & P-n21-k2 & 262.44 & 293.8 & 328.33 & 470.03 & \textbf{30} & \textbf{0} & 261.37 & 315.97 & 398.92 & 374.03 & \textbf{30} & \textbf{0} & \textbf{254.36} & 360.38 & 413.14 & 345.04 & \textbf{30} & \textbf{0} \\
5 & P-n22-k2 & \textbf{234.92} & 285.36 & 332.76 & 685.04 & \textbf{30} & \textbf{0} & 265.65 & 316.45 & 363.37 & 382.05 & \textbf{30} & \textbf{0} & 319.56 & 371.45 & 409.67 & 342.06 & \textbf{30} & \textbf{0} \\
6 & P-n22-k8 & 748.1 & 1844.12 & 2007.05 & 907.1 & 4 & \textbf{0.87} & \textbf{656.32} & \textbf{1795.18} & 2007.05 & 379.09 & \textbf{5} & 1 & 772.35 & 1844.49 & 2007.05 & 342.09 & 4 & 0.97 \\
7 & P-n23-k8 & 1212.67 & 1212.67 & 1212.67 & 529.09 & 0 & 1.13 & 1212.67 & 1212.67 & 1212.67 & 375.13 & 0 & 1.23 & \textbf{587.73} & \textbf{1191.84} & 1212.67 & 345.1 & \textbf{1} & \textbf{1.1} \\
8 & P-n40-k5 & \textbf{601.21} & 753.54 & 825.29 & 3652.12 & \textbf{30} & \textbf{0} & 730.49 & 831.22 & 915.78 & 1923.12 & \textbf{30} & \textbf{0} & 848.45 & 944.27 & 1043.98 & 2379.13 & \textbf{30} & \textbf{0} \\
9 & P-n45-k5 & \textbf{784.5} & 893.73 & 1036.76 & 3863.17 & \textbf{30} & \textbf{0} & 885.76 & 971.29 & 1171.04 & 1990.15 & \textbf{30} & \textbf{0} & 1020.42 & 1120.56 & 1228.87 & 3015.16 & \textbf{30} & \textbf{0} \\
10 & P-n50-k7 & \textbf{830.8} & 986.65 & 1133.58 & 4194.23 & \textbf{30} & \textbf{0} & 935.79 & 1028.14 & 1203.96 & 2137.18 & \textbf{30} & \textbf{0} & 1100.06 & 1203.44 & 1311.18 & 2505.19 & \textbf{30} & \textbf{0} \\
11 & P-n50-k8 & 3478.25 & 3478.25 & 3478.25 & 4061.18 & 0 & 1 & 3478.25 & 3478.25 & 3478.25 & 1911.17 & 0 & 1 & \textbf{1201.58} & \textbf{3402.36} & 3478.25 & 2136.19 & \textbf{1} & \textbf{0.97} \\
12 & P-n50-k10 & \textbf{989.85} & 2256.03 & 3591.08 & 4379.17 & 16 & 0.47 & 1053.89 & \textbf{1814.82} & 3591.08 & 1954.18 & \textbf{22} & \textbf{0.27} & 1193.93 & 1981.23 & 3591.08 & 2189.15 & 21 & 0.3 \\
13 & P-n51-k10 & \textbf{1046.93} & 2888.98 & 3894.33 & 4303.12 & 11 & 0.63 & 1126.53 & \textbf{2480.07} & 3894.33 & 2066.19 & \textbf{16} & \textbf{0.47} & 1304.45 & 3396.3 & 3894.33 & 2205.2 & 6 & 0.8 \\
14 & P-n55-k7 & \textbf{919.34} & 1057.15 & 1176.31 & 4637.16 & \textbf{30} & \textbf{0} & 1003.07 & 1123.59 & 1245.37 & 2105.16 & \textbf{30} & \textbf{0} & 1171.94 & 1312.2 & 1400.57 & 2209.23 & \textbf{30} & \textbf{0} \\
15 & P-n55-k8 & \textbf{947.1} & 1056.65 & 1141.95 & 4606.11 & \textbf{30} & \textbf{0} & 1031.05 & 1157.96 & 1447.38 & 2105.12 & \textbf{30} & \textbf{0} & 1304.47 & 1378.49 & 1467.83 & 2460.19 & \textbf{30} & \textbf{0} \\
16 & P-n55-k10 & \textbf{963.15} & 1135.5 & 1252.9 & 5032.21 & \textbf{30} & \textbf{0} & 1073.89 & 1198.93 & 1526.4 & 1951.19 & \textbf{30} & \textbf{0} & 1296.2 & 1403.19 & 1512.55 & 2691.26 & \textbf{30} & \textbf{0} \\
17 & P-n55-k15 & \textbf{4192.28} & 4192.28 & 4192.28 & 4678.15 & \textbf{0} & 1.27 & \textbf{4192.28} & 4192.28 & 4192.28 & 1854.22 & \textbf{0} & \textbf{1.2} & \textbf{4192.28} & 4192.28 & 4192.28 & 2226.22 & \textbf{0} & 1.23 \\
18 & P-n60-k10 & \textbf{1128.38} & 1569.4 & 4430.1 & 11626.18 & 27 & 0.1 & 1176.77 & 1656.46 & 4430.1 & 6424.18 & 27 & 0.1 & 1466.11 & \textbf{1790.67} & 4430.1 & 6557.24 & \textbf{28} & \textbf{0.07} \\
19 & P-n60-k15 & \textbf{1279.87} & 3102.53 & 4758.57 & 10991.18 & 15 & 0.5 & 1376.12 & 3248.25 & 4758.57 & 6213.19 & 14 & 0.53 & 1649.04 & \textbf{3043.5} & 4758.57 & 6288.18 & \textbf{17} & \textbf{0.43} \\
20 & P-n65-k10 & \textbf{1274.55} & 1499.29 & 5002.05 & 11617.3 & 29 & 0.03 & 1313.05 & \textbf{1471.19} & \textbf{1725.18} & 6696.26 & \textbf{30} & \textbf{0} & 1670.51 & 1777.22 & 1950.48 & 6275.34 & \textbf{30} & \textbf{0} \\
21 & P-n70-k10 & \textbf{1365.95} & 3073.7 & 5449.88 & 12984.3 & 18 & 0.4 & 1473.21 & \textbf{2890.58} & 5449.88 & 6774.24 & 20 & 0.33 & 1770.98 & 2854.5 & 5449.88 & 6405.37 & \textbf{22} & \textbf{0.27} \\
22 & P-n76-k4 & \textbf{1057.05} & 1224.27 & 1359.7 & 47849.37 & \textbf{30} & \textbf{0} & 1193.26 & 1386.96 & 1930.97 & 30126.38 & \textbf{30} & \textbf{0} & 1448.4 & 1796.03 & 2205.91 & 25363.38 & \textbf{30} & \textbf{0} \\
23 & P-n76-k5 & \textbf{1074.07} & 1279.95 & 1466.17 & 54207.32 & \textbf{30} & \textbf{0} & 1301.95 & 1490.24 & 1737.11 & 31396.34 & \textbf{30} & \textbf{0} & 1563.2 & 1859.8 & 2078.12 & 25477.56 & \textbf{30} & \textbf{0} \\
24 & P-n101-k4 & \textbf{1387.68} & 1646.73 & 1891.92 & 66422.27 & \textbf{30} & \textbf{0} & 1610.49 & 1927.66 & 2301.53 & 33917.93 & \textbf{30} & \textbf{0} & 1869.38 & 2532.13 & 3123.3 & 28130.84 & \textbf{30} & \textbf{0} \\		
\bottomrule
\end{tabular}

\begin{flushleft}
\small \textbf{FC:} is the number of feasible solutions obtained out of 30 independent run, and \textbf{$\mathbf{K_{err}}$:} represents the average number of vehicles exceeding the maximum allowed vehicles $K_{max}$.
\end{flushleft}
\end{table}

The results after applying the ML repair heuristic reveal a significant shift in the relative performance of the three encoding strategies. The analysis proceeds along the same dimensions: solution quality, feasibility performance, and fleet violation reduction.

\begin{figure}[ht!]
\centering
\includegraphics[width=\textwidth]{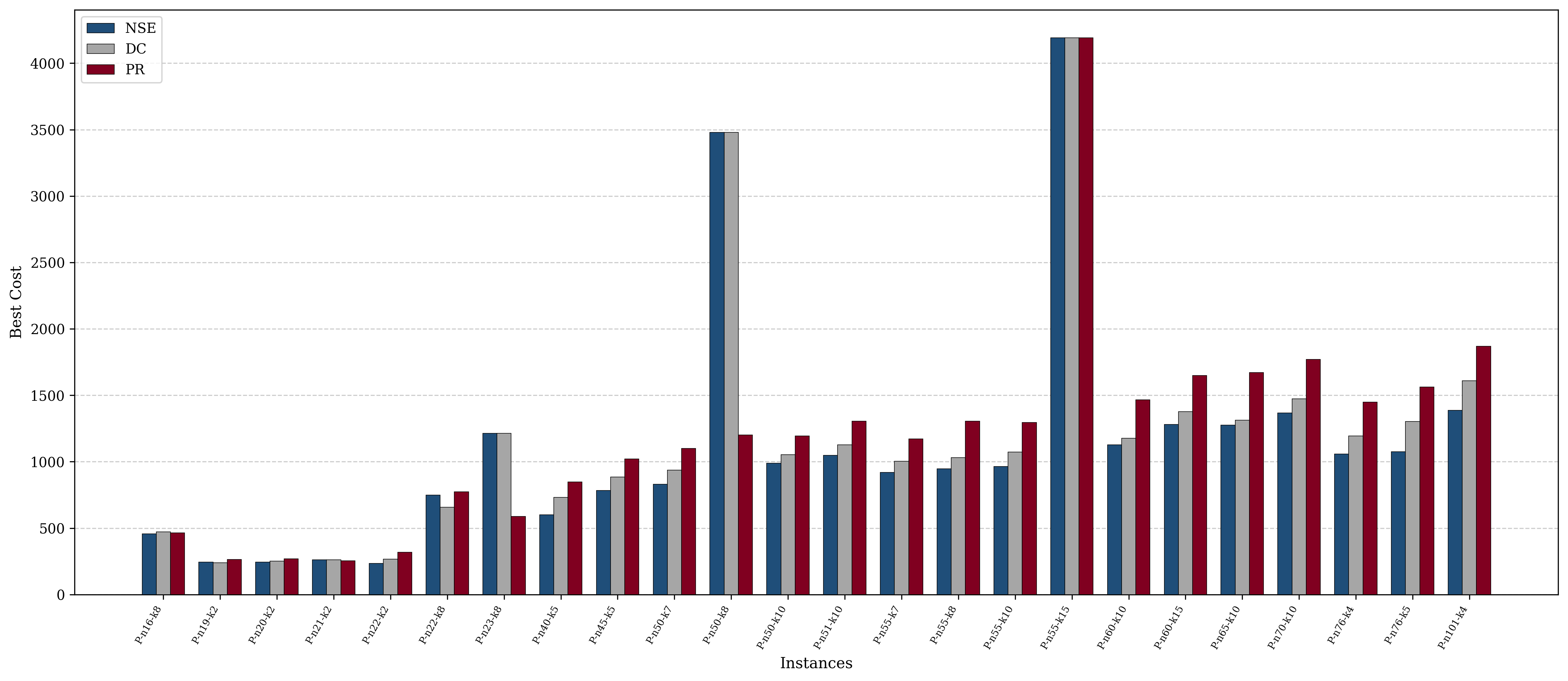}
\caption{Best solution cost comparison across instances after ML heuristic }
\label{fig:feas_comp_MLH}
\end{figure}

Regarding solution quality (Fig.~\ref{fig:feas_comp_MLH}), NSE maintains its dominance. On the 24 benchmark instances, NSE achieves the strictly lowest Best cost on 18 instances, compared to 2 for each PR and DC, with the remaining instance being a tie (P-n23-k8 and P-n55-k15). The same pattern holds for average solution cost: NSE dominates on 14 instances, DC on 5, and PR on only 2. 

\begin{figure}[ht!]
\centering
\includegraphics[width=\textwidth]{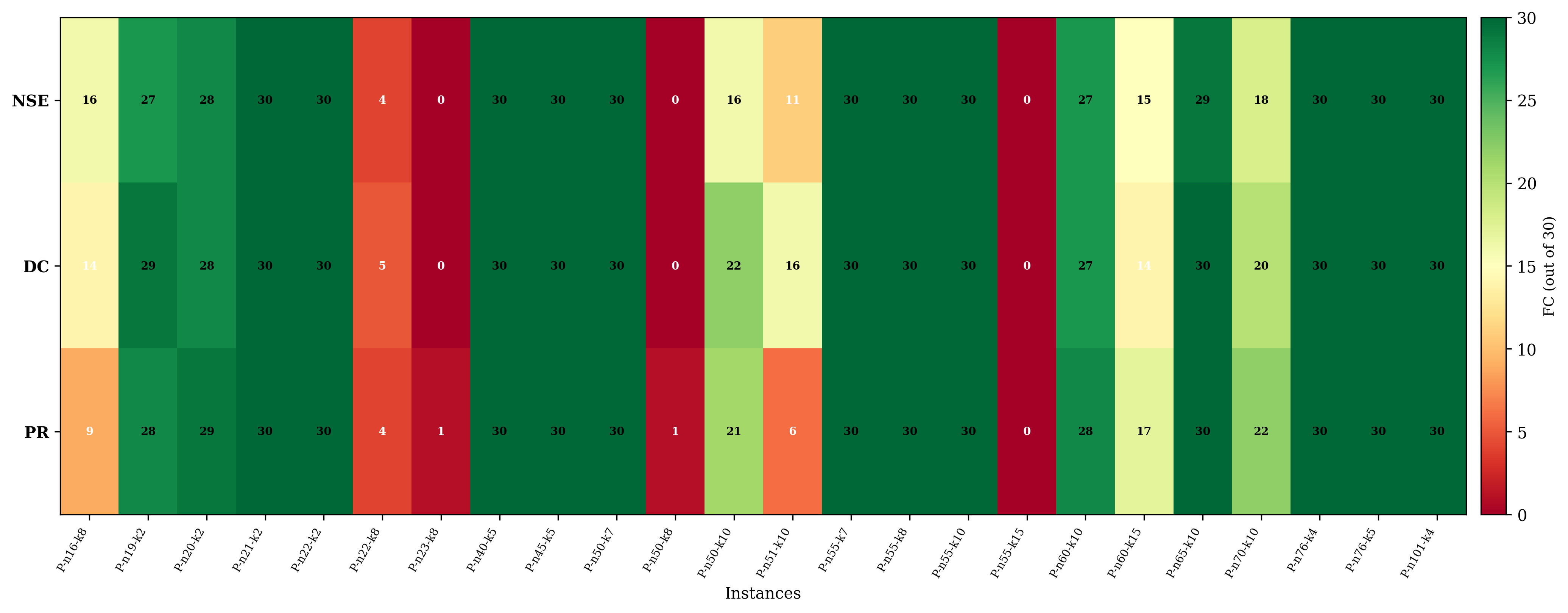}
\caption{Feasibility rate heatmap for NSE, DC, and PR encoding strategies }
\label{fig:feasib_heatmap_MLH}
\end{figure}

The feasibility analysis reveals a striking improvement compared to the pre-repair results (Fig. \ref{fig:feasib_heatmap_MLH}). The number of feasible solutions increases substantially across all encodings; on 14 instances, full feasibility (30 out of 30 runs) is recovered for all three encoding strategies.

\begin{figure}[ht!]
\centering
\includegraphics[width=\textwidth]{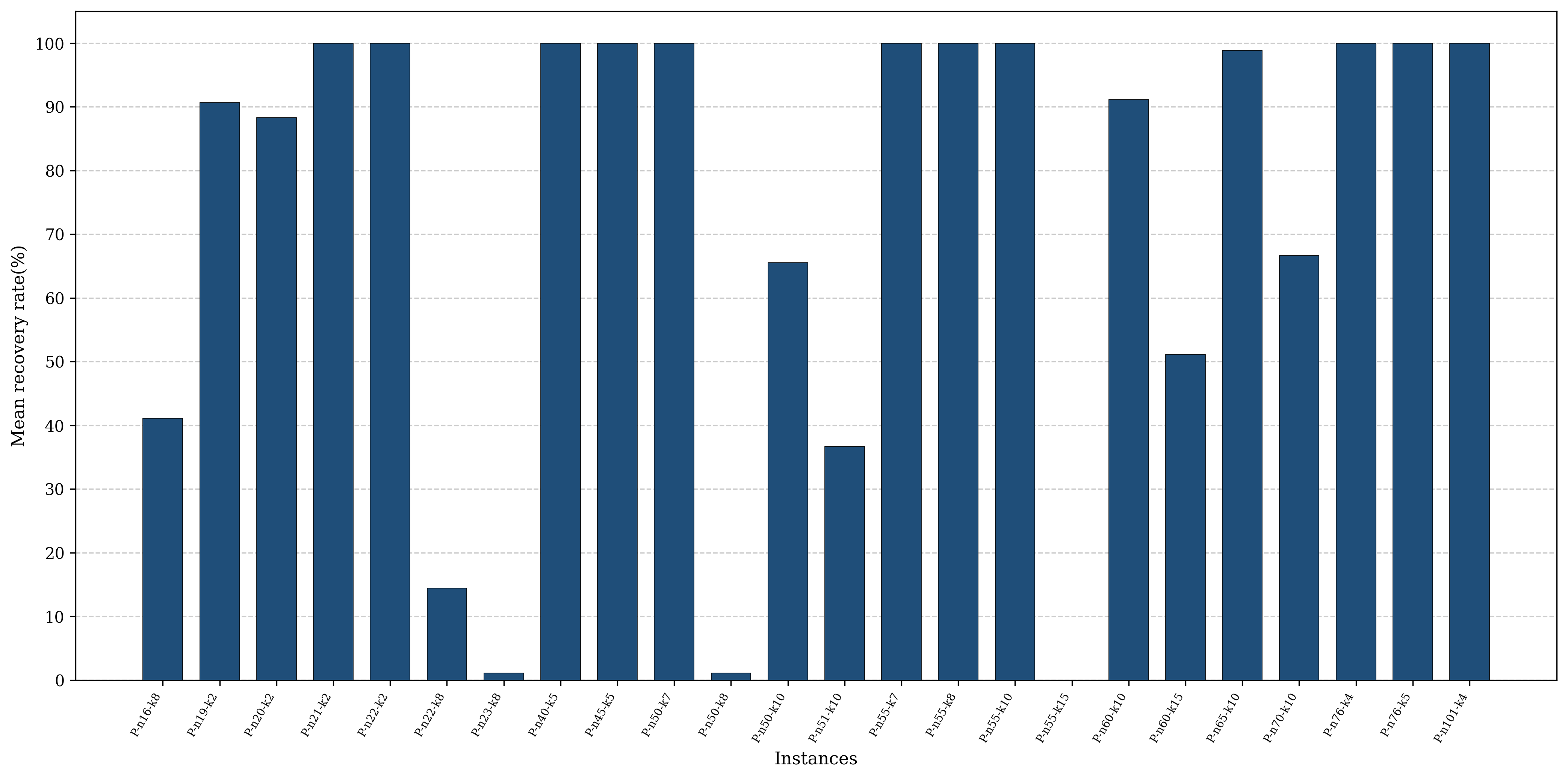}
\caption{Mean recovery rate per instance}
\label{fig:recovery_rate}
\end{figure}

The recovery rate analysis demonstrates the MLH's effectiveness. The overall feasibility recovery rate (total recovered infeasible solutions divided by total infeasible solutions) is 70.46\% for NSE, 72.84 \% for Dc and 71.48\% for PR. High recovery ($>$80\%) is observed on P-n19-k2, P-n20-k2 P-n21-k2, P-n22-k2, P-n40-k5, P-n45-k5, P-n50-k7, P-n55-k7, P-n55-k8, P-n55-k10, P-n60-k10, P-n65-k10, P-n76-k4, P-n76-k5 and P-n101-k4. Moderate recovery (50--80\%) occurs on P-n50-k10, P-n60-k15 and P-n70-k10. Low recovery ($<$50\%) is observed on P-n16-k8, P-n22-k8, P-n23-k8 and P-n50-k8 and P-n51-k10. Zero recovery is recorded on P-n55-k15 (Fig.~\ref{fig:recovery_rate}).

\begin{figure}[ht!]
	\centering
	\includegraphics[width=0.5\textwidth]{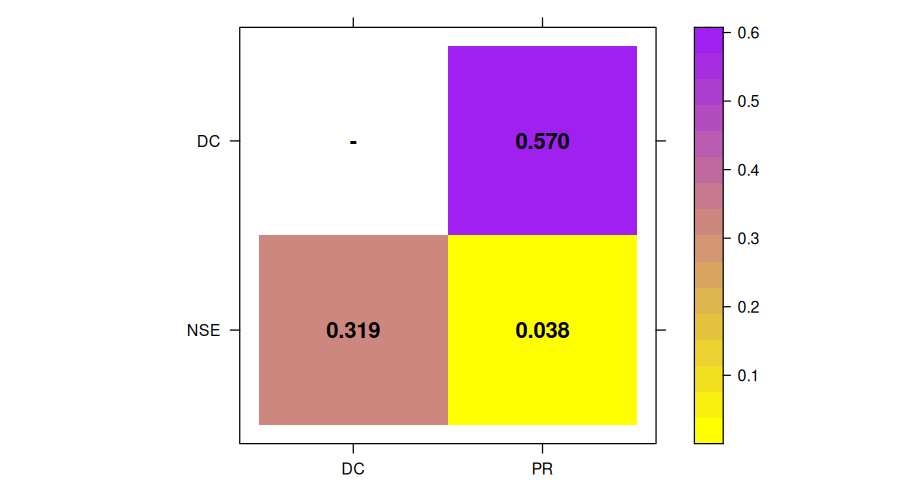}
	\caption{Post-hoc comparison without ML enhancement.}
	\label{posthocWithoutML}
\end{figure}

In order to strengthen the evaluation of the ML heuristic's effect, a statistical ablation analysis was conducted by comparing performance ranks both with and without the ML enhancement. By considering the derived ranks from the best fitness results in Table~\ref{tab:results}, the non-parametric Friedman test~\cite{Friedman} yielded $\chi^2 = 12.978$ ($p = 0.00152$), providing solid evidence to reject the null hypothesis of equal performance. To locate specific pairwise differences, a post-hoc Nemenyi test~\cite{Nemenyi} with single-step $p$-value adjustment was performed. The results, depicted in Figure~\ref{posthocWithoutML}, demonstrate that NSE achieves statistically significant superiority over PR ($p = 0.038$). While NSE also achieved a lower average rank than DC, the difference between them did not reach statistical significance ($p = 0.319$). Similarly, no significant difference was observed between DC and PR ($p = 0.570$).

\begin{figure}[ht!]
	\centering
	\includegraphics[width=0.5\textwidth]{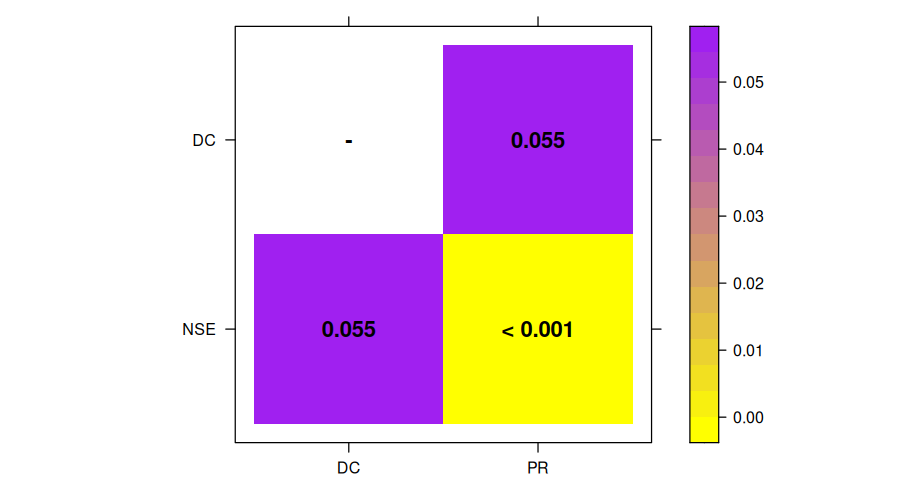}
	\caption{Post-hoc comparison with ML enhancement.}
	\label{posthocWithML}
\end{figure}

Second, the same statistical tests were conducted on the results derived from Table~\ref{tab:results_after_MLH}. With the addition of the ML enhancement, the overall statistical separation increased substantially ($\chi^2 = 22.756$, $p = 1.145 \times 10^{-5}$). The post-hoc comparison also showed a remarkable improvement. As depicted in Figure~\ref{posthocWithML}, the pairwise gap between NSE and PR was strengthened significantly ($p = 1.2 \times 10^{-5}$), while the separation between NSE and DC, as well as between DC and PR, approached standard statistical significance ($p = 0.055$). These results confirm that the ML enhancement consistently boosts solution quality and accentuates performance advantages across test instances.

Finally, we observed that the computational overhead of the ML repair heuristic is marginal. The increase in runtime is very small compared to the time required for the GA and splitting phases, confirming that the heuristic operates as a lightweight post-processing layer rather than a computational bottleneck. It is worth noting that the number of surplus routes observed across the 24 benchmark instances was consistently small, typically two to three routes on average above the fleet limit $K$. This confirms that the theoretical worst-case bound of $O(n^{2})$ is not encountered in practice, and explains why the repair heuristic adds a negligible runtime overhead even on the largest instances.

\subsection{Linear split + ML heuristic vs. Constrained linear split}

Table \ref{tab:comparison_split_strategies} reports the comparative performance of the Linear Split + the proposed ML Heuristic (LSWH) pipeline against the constrained split with limited fleet (SWLF). For each strategy, the table provides: best solution cost (Best), average solution cost (Average), worst solution cost (Worst), average computational time in milliseconds (Time), the number of feasible solutions out of 30 runs (FC), along with lower bounds obtained via GLPK within a 4-hour time limit, and the known optimal solution for each instance.

\begin{table}[ht!]
\centering
\setlength{\tabcolsep}{3.5pt}
\caption{Comparison of Linear Split + MLH and Split with Limited Fleet for NSE Giant tours.}
\label{tab:comparison_split_strategies}

\begin{tabular}{ll lllll lllll ll}
\toprule
\multirow{2}{*}{\textbf{ID}} & \multirow{2}{*}{\textbf{Instance}} 
& \multicolumn{5}{c}{\textbf{Linear Split + MLH}} 
& \multicolumn{5}{c}{\textbf{Split\_with\_limited\_fleet}} 
& \multirow{2}{*}{\textbf{GLPK}} 
& \multirow{2}{*}{\textbf{Optimal}} 
\\
\cmidrule(lr){3-7} \cmidrule(lr){8-12}
& & {Min} & {Mean} & {Max} & {Time (ms)} & {FC} 
  & {Min} & {Mean} & {Max} & {Time (ms)} & {FC} 
  & & \\
\midrule
1 & P-n16-k8 & 456.95 & 653.5 & 856.25 & 677.08 & \textbf{16} & 456.95 & 842.94 & 856.25 & 677.46 & 1 & \textbf{450} & 450 \\
2 & P-n19-k2 & 244.6 & 327.64 & 833.66 & 689.1 & \textbf{27} & 244.6 & 626.14 & 833.66 & 689.2 & 11 & \textbf{212} & 212 \\
3 & P-n20-k2 & 244.69 & 312.97 & 870.56 & 467.05 & \textbf{28} & 244.69 & 415.52 & 870.56 & 467.27 & 23 & \textbf{216} & 216 \\
4 & P-n21-k2 & 262.44 & 293.8 & 328.33 & 470.03 & \textbf{30} & 262.44 & 334.17 & 907.35 & 470.4 & 28 & \textbf{211} & 211 \\
5 & P-n22-k2 & 234.92 & 285.36 & 332.76 & 685.04 & \textbf{30} & 234.92 & 422.71 & 950.22 & 685.26 & 24 & \textbf{216} & 216 \\
6 & P-n22-k8 & 748.1 & 1844.12 & 2007.05 & 907.1 & \textbf{4} & 2007.05 & 2007.05 & 2007.05 & 907.62 & 0 & \textbf{603} & 603 \\
7 & P-n23-k8 & 1212.67 & 1212.67 & 1212.67 & 529.09 & \textbf{0} & 1212.67 & 1212.67 & 1212.67 & 529.64 & \textbf{0} & \textbf{529} & 529 \\
8 & P-n40-k5 & 601.21 & 753.54 & 825.29 & 3652.12 & \textbf{30} & 639.72 & 778.19 & 859.59 & 3652.72 & \textbf{30} & \textbf{574} & 458 \\
9 & P-n45-k5 & 784.5 & 893.73 & 1036.76 & 3863.17 & \textbf{30} & 780.67 & 1088.37 & 3231.73 & 3863.84 & 28 & \textbf{651} & 510 \\
10 & P-n50-k7 & 830.8 & 986.65 & 1133.58 & 4194.23 & \textbf{30} & 867.54 & 1013.59 & 1178.28 & 4195.34 & \textbf{30} & \textbf{743} & 554 \\
11 & P-n50-k8 & \textbf{3478.25} & 3478.25 & 3478.25 & 4061.18 & \textbf{0} & \textbf{3478.25} & 3478.25 & 3478.25 & 4062.31 & \textbf{0} & -- & 631 \\
12 & P-n50-k10 & \textbf{989.85} & 2256.03 & 3591.08 & 4379.17 & \textbf{16} & 3591.08 & 3591.08 & 3591.08 & 4380.65 & 0 & -- & 696 \\
13 & P-n51-k10 & \textbf{1046.93} & 2888.98 & 3894.33 & 4303.12 & \textbf{11} & 3894.33 & 3894.33 & 3894.33 & 4304.75 & 0 & -- & 741 \\
14 & P-n55-k7 & \textbf{919.34} & 1057.15 & 1176.31 & 4637.16 & \textbf{30} & 1006.67 & 1095.66 & 1229.16 & 4638.34 & \textbf{30} & -- & 568 \\
15 & P-n55-k8 & \textbf{947.1} & 1056.65 & 1141.95 & 4606.11 & \textbf{30} & \textbf{947.1} & 1066.29 & 1162.16 & 4607.45 & \textbf{30} & -- & 588 \\
16 & P-n55-k10 & \textbf{963.15} & 1135.5 & 1252.9 & 5032.21 & \textbf{30} & 1083.58 & 1994.18 & 3909.47 & 5033.78 & 21 & -- & 694 \\
17 & P-n55-k15 & \textbf{4192.28} & 4192.28 & 4192.28 & 4678.15 & \textbf{0} & \textbf{4192.28} & 4192.28 & 4192.28 & 4681.13 & \textbf{0} & -- & 989 \\
18 & P-n60-k10 & \textbf{1128.38} & 1569.4 & 4430.1 & 11626.18 & \textbf{27} & 1171.71 & 4006.9 & 4430.1 & 11628.19 & 4 & -- & 744 \\
19 & P-n60-k15 & \textbf{1279.87} & 3102.53 & 4758.57 & 10991.18 & \textbf{15} & 4758.57 & 4758.57 & 4758.57 & 10993.49 & 0 & -- & 968 \\
20 & P-n65-k10 & \textbf{1274.55} & 1499.29 & 5002.05 & 11617.3 & \textbf{29} & 1422.78 & 4410.7 & 5002.05 & 11619.09 & 5 & -- & 792 \\
21 & P-n70-k10 & \textbf{1365.95} & 3073.7 & 5449.88 & 12984.3 & \textbf{18} & 5449.88 & 5449.88 & 5449.88 & 12986.22 & 0 & -- & 827 \\
22 & P-n76-k4 & \textbf{1057.05} & 1224.27 & 1359.7 & 47849.37 & \textbf{30} & 1151.54 & 3481.3 & 5566.67 & 47850.1 & 15 & -- & 593 \\
23 & P-n76-k5 & \textbf{1074.07} & 1279.95 & 1466.17 & 54207.32 & \textbf{30} & 1274.53 & 4095.4 & 5631.3 & 54208.31 & 11 & 1076 & 627 \\
24 & P-n101-k4 & 1387.68 & 1646.73 & 1891.92 & 66422.27 & \textbf{30} & 1681.09 & 1903.04 & 2107.08 & 66422.39 & \textbf{30} & \textbf{1076} & 681 \\
\bottomrule
\end{tabular}

\begin{flushleft}
\small \textbf{FC:} is the number of feasible solutions obtained out of 30 independent runs. -- indicates that GLPK did not find a feasible solution within the time limit.
\end{flushleft}
\end{table}

\begin{figure}[ht!]
\centering
\includegraphics[width=\textwidth]{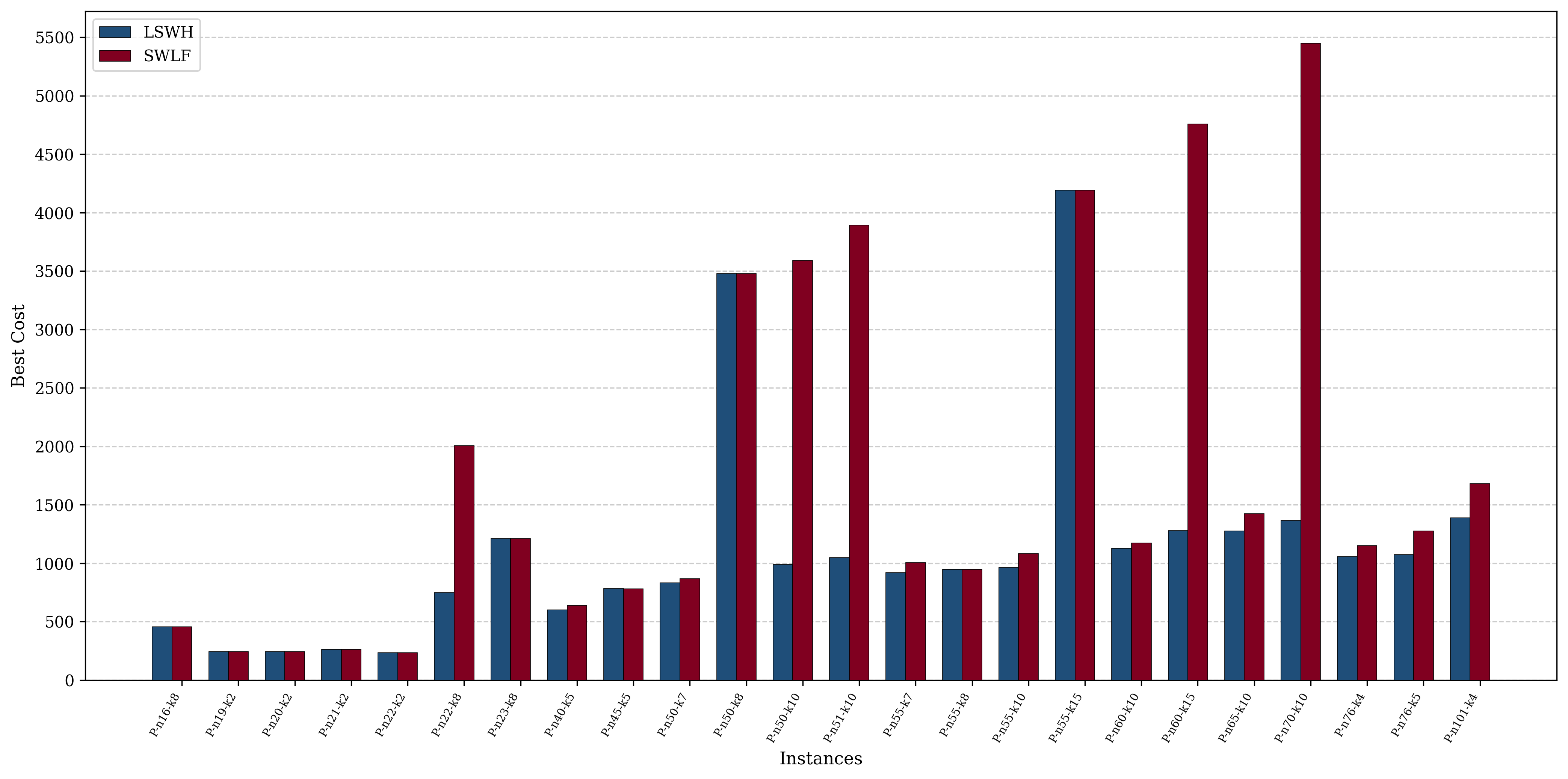}
\caption{Best solution cost comparison between LSWH and SWLF splitting strategies across instances.}
\label{fig:fitness_2_comp}
\end{figure}

The direct comparison between the two splitting strategies leaves little ambiguity. LSWH records a lower best cost on 14 of the 24 instances, SWLF on one (P-n45-k5, where it reaches 780.67 against 784.5), and the two tie on the remaining nine(Fig.\ref{fig:fitness_2_comp}). The mean-cost comparison is even more one-sided: LSWH dominates on 21 instances, with three ties and no wins for SWLF. This asymmetry in the averages is the more important result, because it shows that the ML repair is not merely lucky on a handful of runs but systematically stabilises the search. On P-n76-k4, for instance, the best-cost advantage of LSWH is modest (1057.05 versus 1151.54), yet the mean cost drops from 3481.30 to 1224.27—a reduction of 65\%—because SWLF's infeasible runs are heavily penalised while LSWH's repair heuristic keeps 30 out of 30 runs valid. A similar stabilisation occurs on P-n60-k10 (mean 1569.40 versus 4006.90) and P-n65-k10 (1499.29 versus 4410.70).

\begin{figure}[ht!]
\centering
\includegraphics[width=\textwidth]{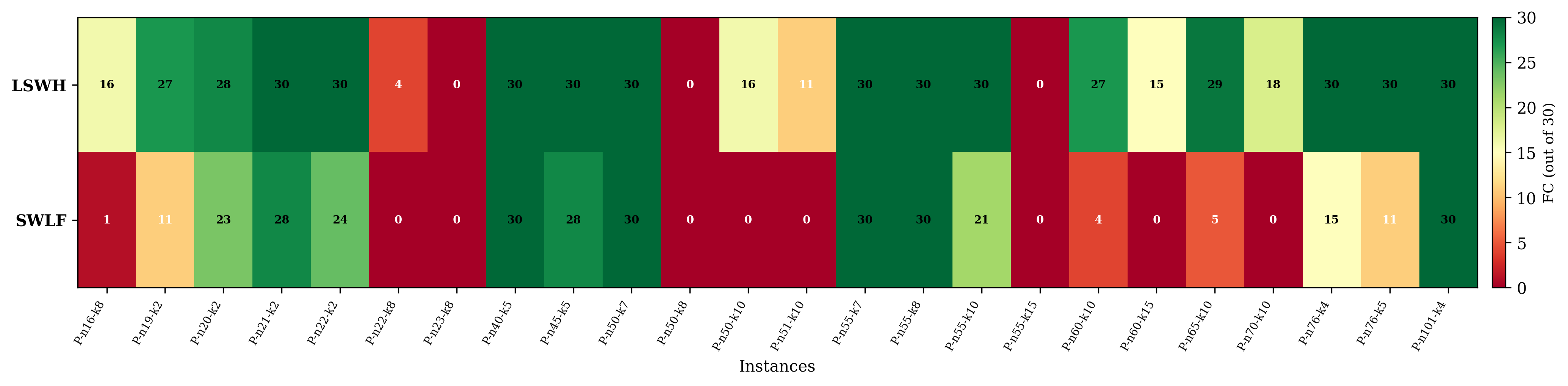}
\caption{Feasibility count heatmap for LSWH and SWLF.}
\label{fig:10}
\end{figure}

\begin{figure}[ht!]
\centering
\includegraphics[width=\textwidth]{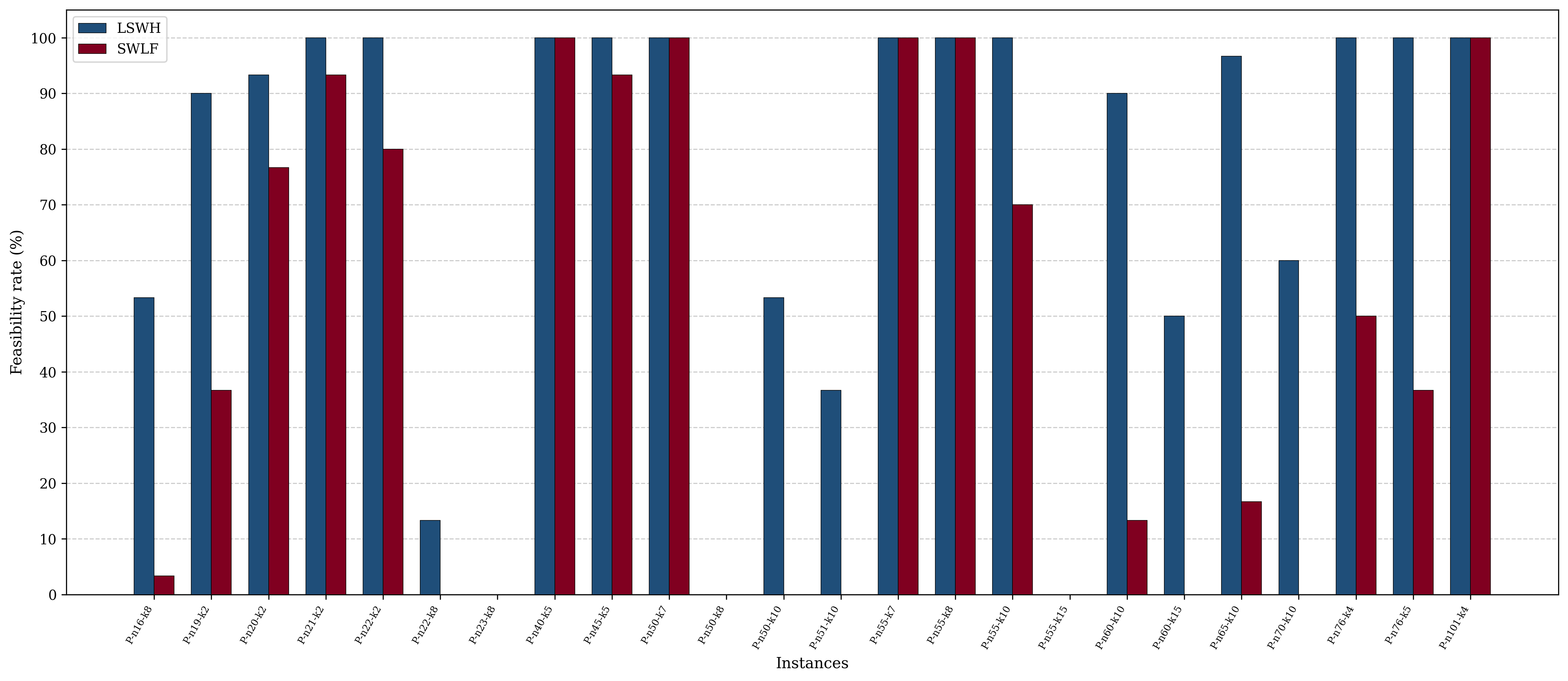}
\caption{feasibility rate per instance.}
\label{fig:11}
\end{figure}

The feasibility counts follow the same pattern (Fig.\ref{fig:10} and Fig.\ref{fig:11}). Across 720 total runs, LSWH produced 521 feasible solutions (72.4\%) against 321 for SWLF (44.6\%). On the small two-vehicle instances P-n19-k2 through P-n22-k2, both methods find identical best minima, but LSWH's feasibility rates are 90\%, 93\%, 100\%, and 100\%, compared with SWLF's 37\%, 77\%, 80\%, and 80\%. On P-n22-k8, LSWH achieves four feasible runs out of 30; SWLF achieves zero. On P-n50-k10, P-n51-k10, and P-n70-k10, LSWH retains feasibility rates of 53\%, 37\%, and 60\%, while SWLF drops to zero on all three. Both methods fail completely (zero feasible runs and identical penalty-dominated costs) only on P23-k8, P-n50-k8 and P-n55-k15. It is worth noting that when the best and average costs coincide, all solutions produced across the 30 runs were infeasible and consequently assigned the same penalty value (see Section~\ref{exp_protocol}). Indeed, without this penalty, the raw fitness of certain infeasible solutions might otherwise misleadingly suggest high solution quality. This penalty mechanism ensures that the reported average fitness remains informative and reliable.

Having established LSWH as the superior pipeline on both best and average metrics, we benchmark it against GLPK and the known optima. On the seven smallest instances (P-n16-k8 through P-n23-k8), GLPK reaches the proven optimal value within four hours on seven of them. LSWH's best costs on these instances range from 1.5\% above optimal (P-n16-k8: 456.95 versus 450) to 24.4\% above (P-n22-k8: 748.1 versus 603).

At medium scale, GLPK begins to struggle. On P-n40-k5, it finds 574 in four hours but does not prove optimality (the known optimum is 458); LSWH reaches 601.21. On P-n45-k5, GLPK delivers 651 against an optimum of 510; LSWH records 784.5. On P-n50-k7, GLPK finds 743 against an optimum of 554; LSWH reaches 830.8. On P-n50-k10 to and P-n55-k10 GLPK returns no feasible solution within four hours, while LSWH achieves best costs.

On P-n76-k5, GLPK obtains 1076 after four hours, while LSWH reaches 1074.07. On P-n101-k4, GLPK again reaches 1076, but LSWH falls to 1387.68. On the remaining large instances—P-n60-k10, P-n60-k15, P-n65-k10, P-n70-k10, and P-n76-k4—GLPK returns no feasible solution within the four-hour limit, whereas LSWH continues to produce feasible tours with best costs of 3478.25, 1046.93, 1128.38, 1279.87, 1274.55, 1365.95, and 1057.05 respectively.

\section{Discussion}
\label{sect:diss}
The results point to a clear conclusion about what works in this sequential framework. The most obvious finding is that Node Shift Encoding consistently outperforms both Path Representation and Double Chromosome. Across the benchmark set, NSE returned better tour costs and produced sequences that were easier to split into a valid number of routes. The reason is straightforward: because NSE manipulates relative shifts rather than absolute positions, the genetic operators tend to preserve useful adjacency patterns between customers. Standard crossover on a direct permutation often tears apart promising sub-sequences, and while Double Chromosome improves on this, it still does not keep the same level of relative ordering intact during decoding. The difference shows up clearly in the $K_{err}$ values. NSE routinely leaves fewer excess routes after the split, which means the repair phase has less work to do and a better chance of succeeding. These observations align with the statistical test results: NSE consistently outperforms PR ($p = 0.038$), while its lead over DC, though evident in average rank, does not reach standard statistical significance ($p = 0.319$). The advantage is therefore genuine, but it is substantially clearer against the baseline Path Representation than against the Double Chromosome encoding.

Turning to the repair heuristic, the Machine Learning post-procedure seems to fill the gap we identified in the classical linear split. When unconstrained splitter returns more than K routes, the ML Heuristic steps in and manages to restore feasibility in approximately 72\% of the evaluated cases. In machine learning heuristic, the centroids are updated continuously after every insertion, so the cluster model learns the new spatial layout of each route on the fly. This dynamic reassignment is essentially an online clustering process, and it works better than a static nearest-centroid rule because it adapts to the changing shape of partial routes.

The comparison with the constrained split is revealing. One might expect that enforcing the fleet limit directly inside the dynamic program (DP) would be the safer approach, but our experiments show otherwise. The constrained split is valid in theory, yet it becomes fragile when the giant tour simply does not contain enough good breakpoints to produce exactly K feasible segments. When that happens, the DP exhausts its state space and falls back on penalty-dominated solutions, which pushes the average cost up dramatically. Our approach takes a different path: we let the split optimize freely for distance, except that we might get too many routes, and then try to fix the fleet size afterward. This pays off. The Linear Split plus ML Heuristic pipeline delivers lower best and average costs and higher feasibility counts on most instances because it does not throw away high-quality partitions that happen to use one or two vehicles too many.

That said, we should be honest about where the method struggles. On a handful of instances, particularly those with very tight capacities and large individual demands, the ML Heuristic fails to recover any feasible solution. In these cases, spatial proximity is simply not enough. The capacity constraints are so strict that the problem resembles a bin-packing task more than a routing task, and no amount of centroid-guided reallocation can fit the customers into K  routes without breaking the load limits. This difficulty is also tied to the ordering of customers in the giant tour. When high-demand customers are poorly sequenced, the routes fill up too quickly, leaving the repair heuristic with little room to maneuver. We also noticed that the decoding time for NSE grows on the largest instances. On P-n101-k4, for example, the framework takes roughly 66 seconds compared to about 28 seconds under Path Representation. However, this increase is modest in absolute terms and, more importantly, it is time well spent. On several medium-to-large instances, GLPK failed to return any feasible solution within a four-hour limit, whereas our pipeline consistently produced valid tours with reasonable costs. When we weigh an extra thirty or forty seconds against the ability to obtain a workable routing plan for problems that exact methods cannot even touch, the trade-off is entirely acceptable. This makes the framework particularly relevant for operational settings where planners need a workable route plan within minutes rather than hours, and where the ability to obtain a reliable solution quickly outweighs the need to prove optimality.

From a practical standpoint, the main advantage of our pipeline is its modularity. The GA can focus on what it does best, which is evolving short Hamiltonian cycles, while the splitting and repair layers handle the operational constraints. This separation makes the code easier to maintain and adapt. If a logistics company later needs to add time windows or heterogeneous fleets, the evolutionary core stays the same; only the insertion rules and capacity checks in the repair phase need adjustment.\\

Despite these encouraging results, there is still ample room to improve the proposed architecture. First, to address instances with poor recovery rates, replacing the random selection of routes to dismantle during repair is an obvious candidate for refinement. Such an improvement must be achieved without adding complexity to the fitness function. Second, adopting adaptive GA parameters could be beneficial. Indeed, crossover and mutation rates could be dynamically adjusted during the search rather than kept static. Finally, the framework currently optimizes total distance alone, but in practice planners often care about route balance. Extending the objective to include driver workload equity would make the approach more attractive for real-world deployment. 

\section{Conclusion and future directions}
\label{sect: conc}
This paper proposes a sequential framework for the CVRP that combines Node Shift Encoding (NSE) with a post-split machine learning repair heuristic within a Giant Tour approach. Computational experiments on well-known benchmarks, supported by statistical evidence, demonstrate that NSE outperforms both Path Representation and Double Chromosome in solution quality and fleet feasibility. The proposed ML heuristic strengthens NSE's superiority by restoring feasibility in approximately 72\% of infeasible cases, while the combined Linear Split and ML heuristic pipeline proves more robust than state-of-the-art literature methods in terms of solution cost and feasibility rate (72.4\% vs 44.6\%). By separating route sequencing from constraint handling, the framework offers a modular and practical approach that is easy to adapt to richer operational settings. Overall, these results suggest that the proposed pipeline is a viable alternative to embedding hard fleet constraints directly within the genetic algorithm. Future work may explore smarter dismantling strategies during repair and multi-objective extensions that balance total distance with route equity.

\section*{Data availability}
The benchmark instances are publicly available from CVRP-LIB.

\section*{Declaration of competing interest}
The authors declare that they have no known competing financial interests or personal relationships that could have appeared to influence the work reported in this paper.

\section*{Funding}
This research received no specific grant from any funding agency in the 
public, commercial, or not-for-profit sectors.

\bibliography{Bib}

\end{document}